%% file: main.tex
\ifdefined\pdftexversion\pdfoutput=1\fi
\documentclass[10pt]{article}
\usepackage[preprint]{tmlr}
\usepackage{hyperref}
\hypersetup{colorlinks=true, linkcolor=[rgb]{0.10,0.27,0.55}, citecolor=[rgb]{0.10,0.27,0.55},
  urlcolor=[rgb]{0.10,0.27,0.55}}
\usepackage{url}
\usepackage{graphicx}
\usepackage{tikz}
\usetikzlibrary{arrows.meta}
\usepackage{booktabs}
\usepackage{multirow}
\usepackage{enumitem}
\usepackage{threeparttable}
\usepackage{caption}
\usepackage{array}
\usepackage{xcolor}
\usepackage{amsmath,amssymb,mathtools,amsthm}
\usepackage[capitalize,noabbrev]{cleveref}
\theoremstyle{plain}
\newtheorem{proposition}{Proposition}
\newif\iflong \longfalse
\newif\ifjmlr \jmlrfalse
\newif\ifcompact \compacttrue  %
\input{shared/macros}
\title{Does a Trading Policy Need to Know What It Trades?\\ Microstructure Invariance and Zero-Shot Transfer\\ of Execution Policies}
\def\month{MM}
\def\year{YYYY}
\author{\name Hieu Tran \email hieutran@umass.edu\\
  \addr University of Massachusetts Amherst
  \AND
  \name Hong Yu\\
  \addr University of Massachusetts Amherst\\
  University of Massachusetts Lowell}
\begin{document}
\maketitle
\input{abstract}
\input{body}

\section*{Broader Impact Statement}
\input{shared/impact}

\input{main.bbl}
\bibliographystyle{tmlr}
\input{appendix}

\end{document}

%% file: shared/macros.tex
\newif\ifshortmain \iflong\ifcompact\shortmaintrue\else\shortmainfalse\fi\else\shortmaintrue\fi
\newif\ificml \iflong\icmlfalse\else\ifcompact\icmlfalse\else\icmltrue\fi\fi  %
\newif\ifshowcore \showcoretrue
\newif\ifshowextra \showextratrue
\newif\ifsupp \suppfalse
\newif\ifshowhistory \showhistorytrue  %
\crefname{supp}{Supplement}{Supplements}
\newcommand{\Rraw}{\textsf{raw}}
\newcommand{\Rrawd}{\textsf{raw+desc}}
\newcommand{\Rz}{\textsf{z-score}}
\newcommand{\Rko}{\textsf{KO}}
\newcommand{\Rms}{\textsf{MS}}
\newcommand{\Rmss}{\textsf{MS}$_\sigma$}
\newcommand{\Rmsd}{\textsf{MS}$_\mathsf{d}$}
\newcommand{\Rmst}{\textsf{MS+tick}}
\newcommand{\Rkot}{\textsf{KO+tick}}
\newcommand{\Rratio}{\textsf{ratio}}
\newcommand{\Rrawr}{\textsf{raw}$^\mathsf{s}$}
\newcommand{\Rrawi}{\textsf{raw+info}}
\newcommand{\Rdimless}{\textsf{dimless}}

\newcommand{\Dhat}{\widehat{D}}

\newcommand{\E}{\mathbb{E}}

\newcommand{\safegraphic}[2][]{\IfFileExists{#2}{\includegraphics[#1]{#2}}{\fbox{\parbox{0.9\linewidth}{\centering\small
  figure pending: \texttt{\detokenize{#2}}}}}}
\InputIfFileExists{numbers.tex}{}{}

\InputIfFileExists{tab_notes.tex}{}{}
\newcommand{\CodeURL}{\url{https://anonymous.4open.science/r/invex-6E30}}

%% file: numbers.tex
\newcommand{\AnalysedIexTest}{688{,}404}
\newcommand{\AnalysedIexTestStocks}{790}
\newcommand{\AnalysedIexTestWide}{831{,}443}
\newcommand{\AnalysedIexTestWideStocks}{1{,}671}
\newcommand{\BootChecked}{12}
\newcommand{\BootKept}{9}
\newcommand{\BootKeptFull}{5}
\newcommand{\BootRefits}{20}
\newcommand{\BootStageOneHOneFullWideHi}{+0.005}
\newcommand{\BootStageOneHOneFullWideLo}{-0.007}
\newcommand{\BootStageTwoHTwoFullWideHi}{+0.015}
\newcommand{\BootStageTwoHTwoFullWideLo}{-0.0002}
\newcommand{\BootValCostUnitShare}{22\%}
\newcommand{\BootValCostUnitShareSeed}{65\%}
\newcommand{\BootValHEightSd}{0.0066}
\newcommand{\BootValHEightSeedSd}{0.0040}
\newcommand{\BootValHOneSd}{0.0038}
\newcommand{\BootValHOneSeedSd}{0.0035}
\newcommand{\BootValHSixSd}{0.0045}
\newcommand{\BootValHSixSeedSd}{0.0048}
\newcommand{\BootValHThreeSd}{0.0031}
\newcommand{\BootValHThreeSeedSd}{0.0025}
\newcommand{\BootValHTwoSd}{0.0046}
\newcommand{\BootValHTwoSeedSd}{0.0042}
\newcommand{\CarlBoxHi}{3/2}
\newcommand{\CarlBoxLo}{2/3}
\newcommand{\CarlCheckBoxSolvable}{0.98}
\newcommand{\CarlDefaultDelta}{0.194}
\newcommand{\CarlDefaultUbar}{0.40}
\newcommand{\CarlDefaultWmax}{2.07}
\newcommand{\CarlEpisodes}{16}
\newcommand{\CarlExtTauMax}{0.748}
\newcommand{\CarlExtTauMin}{0.519}
\newcommand{\CarlFitsNn}{90}
\newcommand{\CarlFitsTr}{90}
\newcommand{\CarlGroup}{8}
\newcommand{\CarlHorizon}{200}
\newcommand{\CarlNnExtNatcVsNat}{+0.523}
\newcommand{\CarlNnExtRawcVsNatc}{+2.144}
\newcommand{\CarlNnExtRawiVsNat}{+2.554}
\newcommand{\CarlNnExtSolvedNatc}{0.000}
\newcommand{\CarlNnInCostNatc}{0.915}
\newcommand{\CarlNnInNatVsNatc}{+0.360}
\newcommand{\CarlNnInNatcVsNat}{-0.360}
\newcommand{\CarlNnInNatcVsNatHi}{-0.161}
\newcommand{\CarlNnInNatcVsNatLo}{-0.605}
\newcommand{\CarlNnInNatcVsNatSeedMax}{-0.240}
\newcommand{\CarlNnInNatcVsNatSeedMin}{-0.506}
\newcommand{\CarlNnInRawVsNatc}{+1.130}
\newcommand{\CarlNnInRawcVsNatc}{-0.018}
\newcommand{\CarlNnInRawcVsNatcHi}{+0.007}
\newcommand{\CarlNnInRawcVsNatcLo}{-0.040}
\newcommand{\CarlNnInRawiVsNat}{-0.089}
\newcommand{\CarlNnInSolvedNatc}{0.622}
\newcommand{\CarlNnInTgNat}{-0.020}
\newcommand{\CarlNnInTgNatc}{+0.008}
\newcommand{\CarlNnInTgNatcFitMax}{+0.069}
\newcommand{\CarlNnInTgNatcFitMin}{-0.040}
\newcommand{\CarlNnInTgRaw}{+0.283}
\newcommand{\CarlNnInTgRawHi}{+0.530}
\newcommand{\CarlNnInTgRawLo}{+0.055}
\newcommand{\CarlNnInTgShareRawSeedMax}{0.39}
\newcommand{\CarlNnInTgShareRawSeedMin}{0.08}
\newcommand{\CarlNnSimNatcVsNat}{-0.446}
\newcommand{\CarlNnSimRawVsNatc}{+0.865}
\newcommand{\CarlNnSimRawcVsNatc}{+1.545}
\newcommand{\CarlNnSimRawcVsNatcHi}{+3.077}
\newcommand{\CarlNnSimRawcVsNatcLo}{+0.411}
\newcommand{\CarlNnSimRawiVsNat}{-0.098}
\newcommand{\CarlNnSimTgRaw}{+0.025}
\newcommand{\CarlNnSimTgRawc}{+1.579}
\newcommand{\CarlOracleInCost}{0.779}
\newcommand{\CarlOutFactorHi}{2}
\newcommand{\CarlOutFactorLo}{1.25}
\newcommand{\CarlSeeds}{3}
\newcommand{\CarlSimContexts}{8}
\newcommand{\CarlStates}{576}
\newcommand{\CarlTestContexts}{32}
\newcommand{\CarlTrExtNatVsOracle}{+0.039}
\newcommand{\CarlTrExtNatcVsNat}{+0.072}
\newcommand{\CarlTrInCostNatc}{0.908}
\newcommand{\CarlTrInNatVsOracle}{+0.034}
\newcommand{\CarlTrInNatcVsNat}{+0.094}
\newcommand{\CarlTrInNatcVsNatHi}{+0.133}
\newcommand{\CarlTrInNatcVsNatLo}{+0.061}
\newcommand{\CarlTrInNatcVsNatSeedMax}{+0.124}
\newcommand{\CarlTrInNatcVsNatSeedMin}{+0.071}
\newcommand{\CarlTrInRawVsNatc}{+0.464}
\newcommand{\CarlTrInRawVsNatcHi}{+0.727}
\newcommand{\CarlTrInRawVsNatcLo}{+0.238}
\newcommand{\CarlTrInRawcVsNatc}{+0.699}
\newcommand{\CarlTrInRawiVsNat}{+0.364}
\newcommand{\CarlTrInSolvedNatc}{0.976}
\newcommand{\CarlTrInTgNatcFitMax}{+0.083}
\newcommand{\CarlTrInTgNatcFitMin}{-0.121}
\newcommand{\CarlTrInTgRaw}{+0.023}
\newcommand{\CarlTrInTgRawHi}{+0.066}
\newcommand{\CarlTrInTgRawLo}{-0.021}
\newcommand{\CarlTrInTgRawc}{+0.061}
\newcommand{\CarlTrInTgShareRawSeedMax}{0.14}
\newcommand{\CarlTrInTgShareRawSeedMin}{-0.04}
\newcommand{\CarlTrSimTgRaw}{-0.526}
\newcommand{\CarlTrSimTgRawHi}{+0.126}
\newcommand{\CarlTrSimTgRawLo}{-1.222}
\newcommand{\CarlTrSimTgRawc}{+0.536}
\newcommand{\CarlTrainContexts}{64}
\newcommand{\CarlTrainTauMax}{0.377}
\newcommand{\CarlTrainTauMin}{0.176}
\newcommand{\CollapseAssetDays}{1{,}189}
\newcommand{\CryptoPoolCOneAll}{-0.234}
\newcommand{\CryptoPoolCThreeAll}{+0.059}
\newcommand{\CryptoPoolCTwoAll}{-0.011}
\newcommand{\CryptoPoolCTwoAllHi}{+0.049}
\newcommand{\CryptoPoolCTwoAllLo}{-0.067}
\newcommand{\CryptoPoolDays}{34}
\newcommand{\CryptoProspNovCTwoAll}{-0.189}
\newcommand{\CryptoQaCrossed}{0}
\newcommand{\CryptoQaDays}{49}
\newcommand{\CryptoQaExecutions}{6{,}244{,}770}
\newcommand{\CryptoQaFifoBad}{0}
\newcommand{\CryptoQaMessages}{869 million}
\newcommand{\CryptoQaOutageMinutes}{17}
\newcommand{\CryptoQaPairDays}{759}
\newcommand{\CryptoQaPairs}{16}
\newcommand{\CryptoQaReseeds}{99}
\newcommand{\CryptoRepCTwoAll}{-0.033}
\newcommand{\CryptoRepCTwoAllHi}{+0.029}
\newcommand{\CryptoRepCTwoAllLo}{-0.092}
\newcommand{\CryptoTestCTwoAll}{+0.027}
\newcommand{\CryptoTestCTwoAllHi}{+0.138}
\newcommand{\CryptoTestCTwoAllLo}{-0.077}
\newcommand{\CryptoXZeroVsRef}{-0.131}
\newcommand{\CtmpStageOneMstickYears}{+0.003}
\newcommand{\CtmpStageOneRawVsMstick}{+0.025}
\newcommand{\CtmpStageTwoMstickYears}{+0.007}
\newcommand{\CtmpStageTwoMstickYearsHi}{+0.008}
\newcommand{\CtmpStageTwoMstickYearsLo}{+0.006}
\newcommand{\CtmpStageTwoRawVsMstick}{+0.015}
\newcommand{\CtxVsInfo}{+0.009}
\newcommand{\CtxVsRawr}{-0.003}
\newcommand{\CurveInfoInfoOneK}{0.524}
\newcommand{\CurveInfoInfoThirtyK}{0.474}
\newcommand{\CurveMsTickOneK}{0.500}
\newcommand{\CurveMsTickTenK}{0.478}
\newcommand{\CurveMsTickThirtyK}{0.477}
\newcommand{\CurveRawHundredK}{0.478}
\newcommand{\CurveSeeds}{3}
\newcommand{\DataCrossedRows}{55}
\newcommand{\DataDays}{19}
\newcommand{\DataDaysPassed}{19}
\newcommand{\DataFifoMax}{99.94\%}
\newcommand{\DataFifoMin}{99.62\%}
\newcommand{\DataMessages}{11 billion}
\newcommand{\DataRows}{13 billion}
\newcommand{\DataSymbolsMax}{12{,}809}
\newcommand{\DataSymbolsMin}{8{,}712}
\newcommand{\DataUnresolved}{0}
\newcommand{\DhatLevelRThree}{0.14}
\newcommand{\DhatLevelRTwo}{0.28}
\newcommand{\DhatLevelRZero}{0.82}
\newcommand{\DistStageOneHOneCvarAbs}{0.064}
\newcommand{\DistStageOneHOneMed}{-0.001}
\newcommand{\DistStageOneHTwoCvar}{-0.432}
\newcommand{\DistStageOneHTwoMean}{+0.017}
\newcommand{\DistStageOneHTwoMed}{-0.021}
\newcommand{\DistStageOneHTwoMedHi}{-0.018}
\newcommand{\DistStageOneHTwoMedLo}{-0.024}
\newcommand{\DistStageTwoHOneCvarAbs}{0.090}
\newcommand{\DistStageTwoHOneMed}{-0.003}
\newcommand{\DistStageTwoHTwoCvar}{-0.481}
\newcommand{\DistStageTwoHTwoMean}{+0.008}
\newcommand{\DistStageTwoHTwoMed}{-0.017}
\newcommand{\EpisodesStageOne}{314{,}133}
\newcommand{\EpisodesStageOneStocks}{1{,}964}
\newcommand{\EpisodesStageTwo}{1{,}256{,}793}
\newcommand{\EpisodesStageTwoStocks}{2{,}629}
\newcommand{\FillExecutedShare}{9.8\%}
\newcommand{\FillOrders}{118{,}993}
\newcommand{\FillStockDays}{397}
\newcommand{\FillTwinBehindFirst}{0.0\%}
\newcommand{\FillTwinMatch}{98.7\%}
\newcommand{\FtTestFtVsZeroEpisode}{-0.011}
\newcommand{\IexDiagCostZero}{107.6}
\newcommand{\IexDiagImmediate}{0.33}
\newcommand{\IexDiagQ}{300}
\newcommand{\IexDiagSide}{sell}
\newcommand{\IexDiagSpreadTicks}{1.5}
\newcommand{\IexDiagStock}{NHP}
\newcommand{\IexDiagWeight}{871}
\newcommand{\IexNoNhpITwo}{-0.029}
\newcommand{\IexNoNhpITwoHi}{+0.011}
\newcommand{\IexNoNhpITwoLo}{-0.070}
\newcommand{\IexOodCtxVsMstick}{+0.110}
\newcommand{\IexReIOneShortfallEpisodes}{551}
\newcommand{\IexReITwoEpisodeHiNinetyfive}{-0.001}
\newcommand{\IexReITwoEpisodeLoNinetyfive}{-0.011}
\newcommand{\IexRePFourEpisodeHiNinetyfive}{+0.015}
\newcommand{\IexRePFourEpisodeLoNinetyfive}{+0.007}
\newcommand{\IexRePOneShortfallEpisodes}{458}
\newcommand{\IexReWOneShortfallEpisodes}{87{,}073}
\newcommand{\IexTestEpisodeIOneAll}{-0.187}
\newcommand{\IexTestEpisodeITwoAll}{-0.006}
\newcommand{\IexTestLmEpisodePFourAll}{+0.011}
\newcommand{\IexTestPpoEpisodePOneAll}{-0.216}
\newcommand{\IexTestWideEpisodeWFiveAll}{+0.103}
\newcommand{\IexTestWideEpisodeWFourAll}{+0.198}
\newcommand{\IexWideWFiveAll}{+0.067}
\newcommand{\IexWideWFourAll}{+0.145}
\newcommand{\InfoIexWidePpoInfoVsMstickEpisode}{+0.016}
\newcommand{\InfoIexWideTreesInfoVsMstickEpisode}{+0.070}
\newcommand{\InfoPpoInfoVsMstick}{-0.000}
\newcommand{\InfoPpoInfoVsMstickHi}{+0.004}
\newcommand{\InfoPpoInfoVsMstickLo}{-0.005}
\newcommand{\InfoPpoVsHighRawVsBase}{-0.002}
\newcommand{\InfoPpoVsHighRawVsMstick}{+0.018}
\newcommand{\InfoPpoVsLowRawVsBase}{+0.004}
\newcommand{\InfoPpoVsLowRawVsMstick}{+0.018}
\newcommand{\InfoPpoVsMidMstickVsBase}{-0.005}
\newcommand{\InfoPpoVsMidRawVsBase}{-0.005}
\newcommand{\InfoPpoVsMidRawVsMstick}{+0.015}
\newcommand{\InfoShareStageOneTrees}{81\%}
\newcommand{\InfoStageOneNnInfoVsMstick}{+0.003}
\newcommand{\InfoStageOneNnRawVsMstick}{+0.043}
\newcommand{\InfoStageOneNnRawrVsMstick}{+0.015}
\newcommand{\InfoStageOneTreesInfoVsMstick}{+0.004}
\newcommand{\InfoStageOneTreesRawVsMstick}{+0.023}
\newcommand{\InfoStageOneTreesRawrVsMstick}{+0.014}
\newcommand{\InfoTreesInfoVsMstick}{+0.002}
\newcommand{\InfoTreesRawVsMstick}{+0.026}
\newcommand{\InfoTuneBestRawVsMstick}{+0.024}
\newcommand{\InfoTuneBestRawVsMstickHi}{+0.031}
\newcommand{\InfoTuneBestRawVsMstickLo}{+0.016}
\newcommand{\InfoTuneBothMstickVsBase}{+0.049}
\newcommand{\InfoTuneBothRawVsBase}{+0.026}
\newcommand{\InfoTuneLeavesMstickVsBase}{+0.019}
\newcommand{\InfoTuneLeavesRawVsBase}{-0.002}
\newcommand{\InfoTuneRoundsInfoVsMstick}{+0.012}
\newcommand{\InfoTuneRoundsMstickVsBase}{+0.009}
\newcommand{\InfoTuneRoundsRawVsBase}{+0.001}
\newcommand{\InfoTuneRoundsRawVsBaseHi}{+0.006}
\newcommand{\InfoTuneRoundsRawVsBaseLo}{-0.004}
\newcommand{\InvGdroTenRawrVsPlain}{+0.007}
\newcommand{\InvIrmOneRawrVsPlain}{-0.001}
\newcommand{\InvVrexOneRawrVsPlain}{+0.007}
\newcommand{\NeffIexTest}{28{,}650}
\newcommand{\NeffIexTestShare}{4\%}
\newcommand{\NeffStageOneShare}{74\%}
\newcommand{\NeffStageTwoShare}{64\%}
\newcommand{\OffTickIexDays}{27}
\newcommand{\OffTickIexRows}{4.8~billion}
\newcommand{\OffTickItchDays}{19}
\newcommand{\OffTickItchRows}{12.8~billion}
\newcommand{\PendDamping}{0.1}
\newcommand{\PendDtRatio}{0.1}
\newcommand{\PendEpisodes}{16}
\newcommand{\PendGamma}{0.98}
\newcommand{\PendGridFineUp}{0.031}
\newcommand{\PendGroup}{8}
\newcommand{\PendHorizon}{200}
\newcommand{\PendInRangeCostNat}{0.602}
\newcommand{\PendInRangeOutsideSample}{3}
\newcommand{\PendInRangeRawVsNat}{+0.578}
\newcommand{\PendInRangeRawVsNatHi}{+0.668}
\newcommand{\PendInRangeRawVsNatLo}{+0.490}
\newcommand{\PendInRangeRawdVsNat}{+1.340}
\newcommand{\PendInRangeRawiVsNat}{+0.525}
\newcommand{\PendInRangeRawiVsNatHi}{+0.603}
\newcommand{\PendInRangeRawiVsNatLo}{+0.451}
\newcommand{\PendInRangeRawiVsRawr}{+0.042}
\newcommand{\PendInRangeSystemsTotal}{96}
\newcommand{\PendInRangeTauHighRawVsNat}{+0.137}
\newcommand{\PendInRangeTauLowRawVsNat}{+1.083}
\newcommand{\PendInRangeTauMidRawiVsNat}{+0.204}
\newcommand{\PendInRangeTgNat}{-0.017}
\newcommand{\PendInRangeTgNatFitMax}{+0.011}
\newcommand{\PendInRangeTgNatFitMin}{-0.115}
\newcommand{\PendInRangeTgRaw}{-0.008}
\newcommand{\PendInRangeTgRawHi}{+0.027}
\newcommand{\PendInRangeTgRawLo}{-0.045}
\newcommand{\PendInRangeTgRawd}{+0.008}
\newcommand{\PendIters}{50}
\newcommand{\PendNnAct}{ReLU}
\newcommand{\PendNnEpochs}{10}
\newcommand{\PendNnHidden}{256}
\newcommand{\PendNnInRangeCostNat}{0.733}
\newcommand{\PendNnInRangeRawVsNat}{+0.312}
\newcommand{\PendNnInRangeRawiVsNat}{+0.079}
\newcommand{\PendNnInRangeRawiVsNatHi}{+0.110}
\newcommand{\PendNnInRangeRawiVsNatLo}{+0.052}
\newcommand{\PendNnInRangeRawiVsRawr}{-0.251}
\newcommand{\PendNnInRangeTauLowRawiVsNat}{+0.170}
\newcommand{\PendNnInRangeTgNat}{+0.036}
\newcommand{\PendNnInRangeTgNatFitMax}{+0.104}
\newcommand{\PendNnInRangeTgNatFitMin}{-0.041}
\newcommand{\PendNnInRangeTgRaw}{-0.026}
\newcommand{\PendNnLayers}{2}
\newcommand{\PendNnLr}{0.001}
\newcommand{\PendNnOutPoolTgRaw}{+0.026}
\newcommand{\PendNnOutPoolTgRawi}{+0.430}
\newcommand{\PendNnOutRangeRawiVsNat}{+1.095}
\newcommand{\PendOutPoolTgRaw}{-0.059}
\newcommand{\PendOutPoolTgRawi}{-0.006}
\newcommand{\PendOutRangeCostNat}{0.601}
\newcommand{\PendOutRangeRawVsNat}{+1.289}
\newcommand{\PendOutRangeTgRaw}{-0.022}
\newcommand{\PendOutSystems}{16}
\newcommand{\PendRounds}{200}
\newcommand{\PendSeeds}{3}
\newcommand{\PendStates}{576}
\newcommand{\PendSweepDiffHalf}{+0.188}
\newcommand{\PendSweepDiffHalfHi}{+0.290}
\newcommand{\PendSweepDiffHalfLo}{+0.093}
\newcommand{\PendSweepDiffQuarter}{-0.027}
\newcommand{\PendSweepDiffSixteenth}{-0.044}
\newcommand{\PendSweepDiffSixtyFourth}{-0.035}
\newcommand{\PendSweepDiffSixtyFourthHi}{+0.049}
\newcommand{\PendSweepDiffSixtyFourthLo}{-0.134}
\newcommand{\PendSweepExcessRaw}{+0.614}
\newcommand{\PendSweepHalfMinusSixtyFourthDiff}{+0.223}
\newcommand{\PendSweepHalfMinusSixtyFourthDiffHi}{+0.346}
\newcommand{\PendSweepHalfMinusSixtyFourthDiffLo}{+0.113}
\newcommand{\PendSweepNatExclZero}{1}
\newcommand{\PendSweepNatExclZeroOneLevel}{3}
\newcommand{\PendSweepSeeds}{9}
\newcommand{\PendSweepSystems}{4}
\newcommand{\PendTestSystems}{32}
\newcommand{\PendTrainSystems}{64}
\newcommand{\PendURatio}{0.5}
\newcommand{\PendXfourInRangeNatChange}{+0.005}
\newcommand{\PendXfourInRangeRawiGapChange}{+0.341}
\newcommand{\PendXfourInRangeRawiGapChangeHi}{+0.431}
\newcommand{\PendXfourInRangeRawiGapChangeLo}{+0.245}
\newcommand{\PendXfourStates}{2{,}304}
\newcommand{\PerStockFiveMinus}{+0.085}
\newcommand{\PpoCostUnitShare}{36\%}
\newcommand{\PpoMsTickVsTrees}{-0.029}
\newcommand{\PpoRawVsMsTick}{+0.014}
\newcommand{\PpoStageTwoRawVsMsTick}{+0.014}
\newcommand{\PpoStageTwoVsTreesMsTick}{-0.027}
\newcommand{\PpoTgMsTick}{+0.004}
\newcommand{\PpoTgMsTickHi}{+0.009}
\newcommand{\PpoTgMsTickLo}{-0.0004}
\newcommand{\PpoZVsMsTick}{+0.016}
\newcommand{\ProfReal}{-0.001}
\newcommand{\ProfRealSeedHi}{+0.004}
\newcommand{\ProfRealSeedLo}{-0.006}
\newcommand{\RFourStageOneGrossMsTickVsHistMsTick}{-0.006}
\newcommand{\RFourStageTwoGrossMsTickVsHistMsTick}{-0.006}
\newcommand{\RFourValGrossOffsetsMsTickVsMsTick}{-0.000}
\newcommand{\ReplayBeyondIexAll}{38\%}
\newcommand{\ReplayBeyondIexOnetick}{8\%}
\newcommand{\ReplayBeyondNasdaq}{34\%}
\newcommand{\ReplayBeyondNasdaqMedian}{0.67}
\newcommand{\ReplayShortNasdaqVal}{2}
\newcommand{\ReplayShortNasdaqValEpisodes}{45{,}444}
\newcommand{\RevStageOneArmNnPopVsMsTick}{+0.005}
\newcommand{\RevStageOneArmNnRawVsMs}{+0.031}
\newcommand{\RevStageOneEconRawVsMsBps}{+0.46}
\newcommand{\RevStageOneNnCostUnitShare}{88\%}
\newcommand{\RevStageOneTgSeenRThreetickStocks}{1{,}001}
\newcommand{\RevStageOneTgSeenRZeroAll}{-0.001}
\newcommand{\RevStageOneXabsSpreadGapRThreetickOutside}{+0.019}
\newcommand{\RevStageOneXabsTradesizeGapRThreetickAbove}{+0.017}
\newcommand{\RevStageTwoEconRawVsMsBpsNet}{-0.33}
\newcommand{\RevStageTwoEconRawVsMsBpsNetHi}{-0.14}
\newcommand{\RevStageTwoEconRawVsMsBpsNetLo}{-0.53}
\newcommand{\RevStageTwoTgSeenRThreetickStocks}{737}
\newcommand{\RoblatOneZeromsContrastsRZeroRThreeAll}{+0.018}
\newcommand{\RoboptimisticContrastsRZeroRThreeAll}{+0.012}
\newcommand{\RoboptimisticVsPostThenCrossRThreetickAll}{-0.058}
\newcommand{\RobpessimisticContrastsRZeroRThreeAll}{+0.025}
\newcommand{\RobpessimisticVsPostThenCrossRThreetickAll}{-0.101}
\newcommand{\RoundLotIexTestShare}{8.3\%}
\newcommand{\RoundLotIexTestStocks}{201}
\newcommand{\RoundLotStageTwoMedianPrice}{405}
\newcommand{\RoundLotStageTwoShare}{10.8\%}
\newcommand{\RoundLotStageTwoStocks}{61}
\newcommand{\RulesPerStockMinusPooled}{+0.013}
\newcommand{\SmallOrderLargeHSix}{-0.006}
\newcommand{\SmallOrderLargeHThree}{+0.018}
\newcommand{\SmallOrderLargeHTwo}{+0.022}
\newcommand{\SmallOrderShare}{20\%}
\newcommand{\SmallOrderSmallEpisodes}{9{,}040}
\newcommand{\SmallOrderSmallHSix}{-0.013}
\newcommand{\SmallOrderSmallHThree}{+0.021}
\newcommand{\SmallOrderSmallHTwo}{+0.012}
\newcommand{\SmallOrderSmallHTwoHi}{+0.033}
\newcommand{\SmallOrderSmallHTwoLo}{-0.010}
\newcommand{\StageOneContrastsROneRThreeLarge}{+0.014}
\newcommand{\StageOneContrastsROneRThreeSmall}{+0.004}
\newcommand{\StageOneContrastsRThreetickRThreeLarge}{-0.006}
\newcommand{\StageOneContrastsRThreetickRThreeSmall}{-0.007}
\newcommand{\StageOneContrastsRZeroRThreeLarge}{+0.028}
\newcommand{\StageOneContrastsRZeroRThreeSmall}{+0.008}
\newcommand{\StageOneContrastsRZerodescRThreeLarge}{+0.029}
\newcommand{\StageOneContrastsRZerodescRThreeSmall}{+0.005}
\newcommand{\StageOneContrastsRZerorRThreeLarge}{+0.004}
\newcommand{\StageOneContrastsRZerorRThreeSmall}{+0.013}
\newcommand{\StageOneGainBpsImmediate}{6.17}
\newcommand{\StageOneGainBpsPost}{2.42}
\newcommand{\StageOneHFourAll}{+0.017}
\newcommand{\StageOneHOneAll}{-0.002}
\newcommand{\StageOneHOneLarge}{-0.003}
\newcommand{\StageOneHOneMid}{-0.000}
\newcommand{\StageOneHOneSmall}{-0.003}
\newcommand{\StageOneHThreeAll}{+0.015}
\newcommand{\StageOneHTwoAll}{+0.017}
\newcommand{\StageOneMedianSpreadBps}{20.0}
\newcommand{\StageOneOneTickShare}{21\%}
\newcommand{\StageOnePostCostBps}{17.51}
\newcommand{\StageTwoHOneAll}{-0.001}
\newcommand{\StageTwoMedianSpreadBps}{26.3}
\newcommand{\StageTwoMedianSpreadTicks}{4.7}
\newcommand{\StageTwoOneTickShare}{25\%}
\newcommand{\TdrRawrVsPlain}{+0.028}
\newcommand{\TreesStageTwoRawVsMsTick}{+0.011}
\newcommand{\UniverseTrain}{954}
\newcommand{\ValContrastsRZeroRThreeAll}{+0.018}
\newcommand{\ValPlaceboRThreetickAll}{+0.058}
\newcommand{\WinFifteenMstickRetrainVsReg}{-0.002}
\newcommand{\WinFifteenMstickZeroVsReg}{+0.002}
\newcommand{\WinFifteenRawVsRetrainMstick}{+0.028}
\newcommand{\WinSixtyMstickRetrainVsReg}{+0.002}
\newcommand{\WinSixtyMstickZeroVsReg}{+0.001}
\newcommand{\WinSixtyRawVsRetrainMstick}{+0.024}
\newcommand{\XtrapDescActivityOutsideRatio}{17}
\newcommand{\XtrapDescActivityRangeRatio}{3.0}
\newcommand{\XtrapSpreadRThreetickDhatrho}{+0.02}
\newcommand{\XtrapSpreadRZeroDhatrho}{+0.24}
\newcommand{\ZiAdvPcOneContrastRZero}{+0.077}
\newcommand{\ZiAdvPcOneTgRZero}{+0.012}
\newcommand{\ZiAdvPcOnexTgRTwo}{+0.148}
\newcommand{\ZiAdvPcOnexTgRZero}{+0.166}
\newcommand{\ZiAdvPcThreeContrastRThreed}{-0.002}
\newcommand{\ZiAdvPcThreeContrastRThreedHi}{+0.005}
\newcommand{\ZiAdvPcThreeContrastRThreedLo}{-0.008}
\newcommand{\ZiAdvPcTwoContrastExtRThreetick}{-0.014}
\newcommand{\ZiAdvPcTwoContrastExtRThreetickHi}{-0.001}
\newcommand{\ZiAdvPcTwoContrastExtRThreetickLo}{-0.030}
\newcommand{\ZiAdvPcTwoContrastRZero}{+0.080}

%% file: tab_notes.tex
\providecommand{\TabNoteStageOne}{n/a: no seen policy was trained in \Rrawr{} units. The baselines cost 0.642 (crossing at once) and 0.508 (posting, then crossing).}
\providecommand{\TabNoteStageTwo}{n/a: no seen policy was trained in \Rrawr{} units. The baselines cost 0.665 (crossing at once) and 0.522 (posting, then crossing).}

%% file: abstract.tex
\begin{abstract}
Execution algorithms are calibrated asset by asset, because the queues, trade rates and bid--ask spreads they face
differ by orders of magnitude across assets. For reinforcement learning this is zero-shot generalisation across
contexts of very different scales, and microstructure invariance proposes a normalisation fixed by theory: measured in
each asset's own units of time, size and price, markets look alike. In an exact order-by-order replay of public Nasdaq
data, we train policies once, on 954 stocks in 2019, to choose between waiting in the queue and crossing the spread,
and evaluate them unchanged on 1,964 unseen stocks in 2020--21, 2,629 stocks in 2025--26, crypto pairs and a second
exchange. First, a policy that sees a stock only through its recent trade rate, trade size, spread and price increment
(tick) costs within 0.015 spreads per share (a fifth of what learning saves) of a policy whose training data included
the stock, and the stock's own history does not help it; it comes within a similar margin of policies trained on the
second exchange when every episode counts equally, and within a wider one on crypto. Second, in mean cost before fees,
invariant units beat raw units, z-scores and asset descriptors, but by under a basis point: most of their advantage is
information and a common cost scale that a learner can also take as plain inputs, and none of the representations
tested extrapolates to unseen spread regimes. Third, a near-zero transfer gap to a pooled comparator is weak evidence
that a representation is right: we show that such a comparator detects only a small share of what a poor
representation loses, and confirm it in synthetic markets and on pendulums.
\end{abstract}

%% file: body.tex
\input{sec_intro}
\input{sec_related}
\input{sec_setup}
\input{sec_theory}
\input{sec_design}
\input{sec_results}
\input{sec_discussion}

%% file: sec_intro.tex
\section{Introduction}
\label{sec:intro}

\begin{figure}[t]
  \centering
  \safegraphic[width=\textwidth]{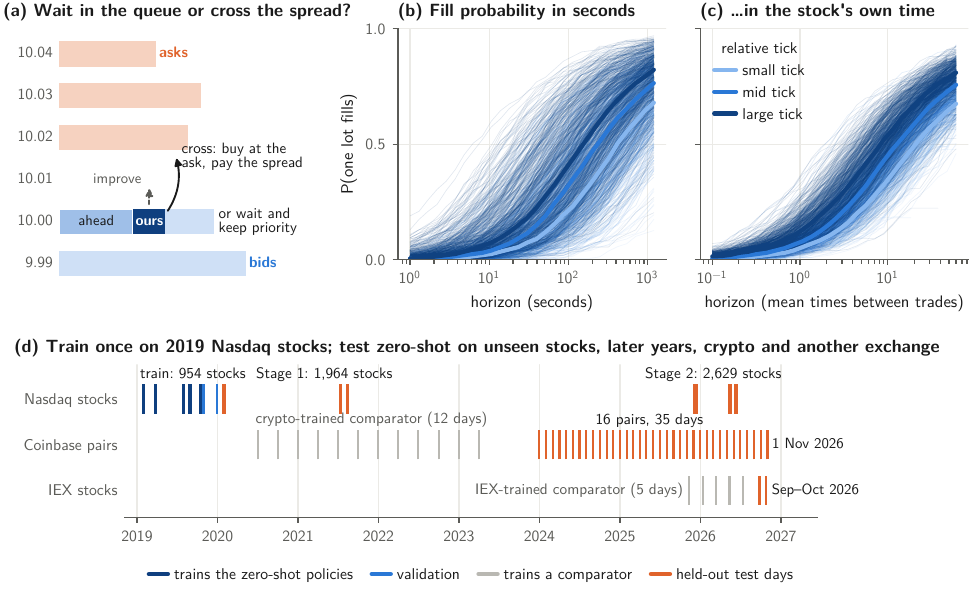}
  \caption{\textbf{One policy for every asset?} (a) The decision we study: a buy order resting at the best bid can
  wait or cross the spread. (b, c) Why units might suffice: the probability that a 100-share order joining the best bid
  fills within a horizon varies widely across \CollapseAssetDays{} training stock-days when time is in seconds, and
  far less in each stock's own mean time between trades (thick lines: medians within terciles of the relative tick,
  the price increment over the volatility per trade). (d) The test:
  policies trained once on 2019 Nasdaq stocks are evaluated unchanged on orange days; comparators trained on
  the test market use the grey days.}
  \label{fig:overview}
\end{figure}

Every execution algorithm keeps facing the same small decision: rest a limit order at the best price and wait in the
queue for a counterparty, or cross the spread, the gap between the best buy and sell prices, by trading now against
the best opposite order (\cref{fig:overview}a). Waiting saves the spread if the order fills and costs more if the price moves
away~\citep{cartea2015algorithmic, bouchaud2018trades}. When to give up depends on the queue ahead, the rate at which
trades arrive, the spread and the volatility, and these differ by orders of magnitude across assets. Execution
policies, hand-tuned or learned, are therefore calibrated asset by asset, and reinforcement learning for execution
and market making has mostly been trained and tested on one asset or a
handful~\citep{nevmyvaka2006rl, spooner2018market, hambly2023reinforcement}. A per-asset policy needs per-asset
data, has nothing to offer a new listing, and does not reveal which properties of an asset it depends on.

Microstructure invariance offers a different starting point: markets look alike once time, size and price are
measured in units set by each asset's own trading. \citet{kyle2016invariance} derive these units from the asset's
price, volume and volatility, \citet{benzaquen2016invariance} from its rate of trades, trade size and spread.
\Cref{fig:overview}b--c shows the idea on our data: the probability that an order at the best price fills within
a horizon varies widely across stocks in seconds, and far less in each stock's own mean time between trades. Both
theories have been tested on statistics of trades and costs. We test them on decisions. For reinforcement learning,
this is zero-shot generalisation across contexts~\citep{kirk2023generalisation}: the assets are contexts whose scales
of time, size and price differ by orders of magnitude, and invariance supplies a normalisation of observations and
rewards, fixed by theory, under which they would become one task.

If two assets are statistically similar in some units, the optimal execution policy written in those units is the
same for both (\cref{prop:similarity}). Real stocks are never exactly similar, so we test the prediction empirically.
We replay every order of the public Nasdaq sample days of 2019--2026 with queue position and latency\footnote{Code: \CodeURL.}, train one
learner on eleven state representations, from raw physical units through standard normalisations to two families
of invariant units (and two diagnostic variants), and evaluate the policies zero-shot on held-out stocks, later years, crypto pairs and a second
exchange, whose last ten test days had not yet occurred when the policies were frozen (\cref{fig:overview}d). We ask:
\begin{enumerate}[label=\textbf{Q\arabic*.}, leftmargin=2.6em, itemsep=1pt, topsep=3pt]
  \item Does a policy trained on some stocks execute unseen stocks, and stocks six years later, as well as a policy
  that saw them? (\cref{sec:res-stocks})
  \item What do invariant units supply that other representations lack? (\cref{sec:res-supply})
  \item Does the transfer reach another market and another exchange? (\cref{sec:res-markets})
  \item Is a small transfer gap evidence that a representation is right? (\cref{sec:res-gap})
\end{enumerate}

\paragraph{Findings.}
\begin{itemize}[leftmargin=1.5em, itemsep=2pt, topsep=3pt]
  \item \textbf{A stock's recent scales and its tick are enough, within the range seen in training.} The zero-shot
  policy costs within 0.015 spreads per share of a pooled policy whose data included the test stock (its
  \emph{transfer gap}), in 2020--21 and in 2025--26. That test alone is weak (Q4); what carries the claim is that the
  stock's own history, or a policy trained on the stock alone, does no better. On a second exchange the policy comes
  within a similar margin when every episode counts equally.
  \item \textbf{The units matter less than theory suggests.} In mean cost before fees, raw units, z-scores and asset
  descriptors lose by about one margin or less, under a basis point; after fees, and at the median, raw units are as
  cheap or cheaper. Most of that loss is a missing cost scale and information: given the units' window statistics as
  inputs and the cost in spreads, raw features recover \InfoShareStageOneTrees{} of it. In a first test on validation
  days, invariance penalties, a context encoder and domain randomisation do no better. The units do save data.
  \item \textbf{Transfer reaches another market, but not unseen regimes.} On crypto pairs the stock-trained policy
  beats crossing the spread and comes within a wider margin of a policy trained on the pairs (over all
  \CryptoPoolDays{} days; the quarterly days alone are too noisy to decide), but none of the representations tested
  carries a policy into a spread regime it never saw.
  \item \textbf{A small transfer gap is weak evidence of invariance.} A pooled comparator that also saw the test
  stock improves on it only in proportion to the stock's small share of its data (\cref{prop:pooled}), so poor units
  show a near-zero gap too; the contrasts between representations carry the evidence. In synthetic markets and in
  pendulums, wrong units cost more without a larger gap.
\end{itemize}

\paragraph{Contributions.}
(1) When a policy should transfer across tasks of different scales: similarity in some units implies an invariant
optimal policy, regret and aliasing bounds say where transfer fails, and a pooled comparator's transfer gap registers
a representation's loss only in proportion to the test task's share of its data (\cref{sec:theory}). (2) A zero-shot
evaluation of eleven representations and three learners across stocks, years, markets and exchanges, with synthetic
controls and pendulums, where the answer is known, and what it says a policy needs to know about the asset it trades
(\cref{sec:design,sec:results,sec:discussion}). (3) An open-source, exact and validated queue-tracking replay of the public Nasdaq sample
days, with IEX and Coinbase parsers (\cref{sec:setup}).

%% file: sec_related.tex
\section{Related work}
\label{sec:related}

\paragraph{Microstructure invariance.}
\citet{kyle2016invariance} hypothesise that bets and their costs are distributed alike across assets in business
time, from dimensional analysis and an invariance principle~\citep{kyle2016dimensional}; the scaling laws hold on US
stocks~\citep{kyle2020invariance} and futures~\citep{andersen2015intraday}. \citet{benzaquen2016invariance} find the
invariant better normalised by the cost of crossing the spread~\citep[see also][]{bucci2020invariants}, and large-tick
assets behave differently~\citep{eisler2012impact, dayri2015tick}. Deep models pool stocks for prediction, where
normalising by volatility or spread did not help~\citep{sirignano2019universal, zhang2019deeplob}. These works test
invariance through statistics and predictions, ours through decisions.

\paragraph{Learning to execute.}
RL for execution~\citep{nevmyvaka2006rl, ning2021double} and market making~\citep{chan2001electronic,
spooner2018market} mostly trains and tests on one asset or a handful~\citep{hambly2023reinforcement}. The closest
work pools assets: one policy across 3{,}566 instruments with rewards normalised by price and order
size~\citep{fang2021universal}, rewards in a stock's trailing half-spread~\citep{moallemi2022rlexecution}, one PPO
policy for 50 stocks~\citep{byun2023practical}, and the market as an uncontrollable context~\citep{zhang2023generalizable}.
We add competing unit hypotheses and the tick, tests on thousands of unseen stocks, later years and other venues, and
controls with planted structure. Replay~\citep{frey2023jaxlob, jaxmarlhft2025} and simulated
books~\citep{amrouni2021abides, cont2010stochastic, huang2015queue} provide environments.

\paragraph{Generalisation and normalisation in RL.}
Zero-shot generalisation is usually framed through contextual MDPs~\citep{hallak2015contextual,
kirk2023generalisation, benjamins2023contextualize}; here the context is the asset and the units an observable
summary of it. Invariance can be learned across environments~\citep{arjovsky2019irm, krueger2021rex,
sagawa2020groupdro} or the context inferred by an encoder~\citep{rakelly2019efficient}; we impose it through units,
as units-equivariant models do~\citep{villar2023dimensionless}, and our similarity proposition is dimensional
analysis~\citep{buckingham1914similar} for policies, an instance of an MDP homomorphism~\citep{ravindran2003smdp,
vanderpol2020mdp}. Units from the target's recent data are test-time normalisation~\citep{schneider2020covariate},
and the reward's scale across tasks is what PopArt learns~\citep{vanhasselt2016popart}. Exogenous
paths~\citep{mao2019variance} let every action be branched on the same path, and our evaluation follows RL's
statistical practice~\citep{henderson2018matters, agarwal2021precipice}.

%% file: sec_setup.tex
\section{The execution problem and its replay}
\label{sec:setup}

\paragraph{The make-or-take problem.}
An exchange's limit order book lists the resting buy and sell orders of an asset. The best bid and the best ask (the
\emph{touch}) are separated by the \emph{spread}, prices lie on a grid whose step is the \emph{tick}, and orders at
one price fill in time priority, so a new order joins the back of a queue. An episode asks the agent to buy (or sell) $Q$ shares within a horizon $T$. At $K = 10$ equally spaced times the agent
can \textsc{hold} its resting order and keep its queue priority, \textsc{join} the near touch (the best price on its own side), \textsc{improve} by
resting one tick inside a spread of at least two ticks, \textsc{cancel}, or \textsc{cross} by taking the remaining
shares from the visible book; whatever is left at $T$ is crossed. Actions reach the exchange after a latency of
$100\,\mu$s ($450\,\mu$s on IEX, whose speed bump adds $350\,\mu$s). The cost per share is $s\,(\bar p - m_0)$, where
$s = \pm 1$ is the side, $\bar p$ the average execution price and $m_0$ the midpoint of the best bid and ask at arrival, reported in units of the
stock's mean spread and in basis points (bp). The agent faces a finite-horizon decision problem: it observes the
order's status and the market features below, a summary of the book, so the problem is partially observed; the
transition is the replayed market plus the agent's own order, and the reward, undiscounted, is minus the cost incurred
until the next decision. \Cref{app:glossary} defines the market terms.

\paragraph{Exact queue replay.}
We replay Nasdaq TotalView-ITCH 5.0 data~\citep{nasdaqitch}, every order added, executed, cancelled or replaced, with
nanosecond timestamps. A virtual limit order arriving at event $i$ queues behind every displayed share at its price;
executions and cancellations of orders created before $i$ move it forward, and once nothing is ahead, the next
executions at its price fill it. Two rules bound this one: a pessimistic rule fills only when the market trades
through the order's price or an opposite order crosses it, an optimistic one as if the order were at the front of its
queue. Because each event records the event that created its order, queue position follows
from integer comparisons, without rebuilding the book. Every stock-day passes book-consistency checks, a fast tracker
agrees fill for fill with an independent one, and a virtual twin at a real order's queue position starts filling when
the real order does in \FillTwinMatch{} of \FillOrders{} cases (\cref{app:data}). The replay is non-intrusive and
covers one venue's displayed book, so crossing a thin book costs more than it would for a routed order; we flag where
this matters.

\paragraph{Representations.}
A representation turns the raw state into features and the cost into a reward. The raw state holds the spread, the
near- and far-touch sizes and their imbalance, the adverse price move since arrival, signed aggressive flow and the
trade count over a lookback, the time since the last trade, and the order's status, shares ahead, time and shares
left, $Q$ and $T$. Most representations divide each quantity by a unit computed from the 30 minutes before the
episode, so \emph{zero-shot} means no training on the target stock, while the policy still reads 30 minutes of its
market data, a form of test-time normalisation~\citep{schneider2020covariate}. The policy is trained on the cost in
the representation's price unit unless noted. We compare three families (\cref{tab:units}):
\begin{itemize}[leftmargin=1.2em, itemsep=2pt, topsep=3pt]
  \item \textbf{Invariant units.} Microstructure units, \Rms~\citep{benzaquen2016invariance}, measure time, size and
  price in the stock's mean time between trades, trade size and spread; Kyle and Obizhaeva's business time,
  \Rko~\citep{kyle2016invariance}, derives the three units from the stock's price, volume and volatility. \Rmst{}
  and \Rkot{} add the tick relative to the price unit and to the volatility per trade, which invariance theory
  leaves out; \Rmss{} measures price in the volatility per trade, and \Rmsd{} size in the touch depth.
  \item \textbf{Raw units.} \Rraw{} keeps seconds, 100-share lots and cents, and \Rrawd{} adds the stock's price,
  volume rate, volatility and tick as inputs, for the network to learn its own normalisation, as contextual RL
  appends the context to the observation~\citep{benjamins2023contextualize}. \Rrawr{} keeps the raw
  features but measures the cost in the stock's spread, separating the features from the scale of the reward;
  \Rrawi, added after the main evaluation, also adds the window statistics the units are built from.
  \item \textbf{Standard normalisations.} \Rz{} standardises each feature over the window (per-context observation
  normalisation), \Rratio{} uses the
  current state's scales, and \Rdimless{} keeps only unitless features; all three measure the cost in spreads.
\end{itemize}

\begin{table}[t]
\centering
\small
\caption{\textbf{Units of each representation.} Features with dimensions are divided by the time, size and price
units, and the policy is trained on the cost per share in the cost unit. $\bar\tau$, $\bar q$, $\bar s$ and $\bar d$
are the mean time between trades, trade size, spread and touch depth over the 30 minutes before the episode,
$\sigma_1$ is the volatility per trade, and $t_\mathrm{KO}$, $q_\mathrm{KO}$ and $p_\mathrm{KO}$ are Kyle and
Obizhaeva's units (\cref{app:glossary}). Tick ratios: the tick over the price unit and over $\sigma_1$, in logs.}
\label{tab:units}
\begin{tabular}{@{}llllll@{}}
\toprule
Representation & Time & Size & Price & Cost & Inputs added \\
\midrule
\multicolumn{6}{@{}l}{\emph{Invariant units}} \\
\quad\Rms & $\bar\tau$ & $\bar q$ & $\bar s$ & $\bar s$ & none \\
\quad\Rmst & $\bar\tau$ & $\bar q$ & $\bar s$ & $\bar s$ & tick ratios \\
\quad\Rmss & $\bar\tau$ & $\bar q$ & $\sigma_1$ & $\sigma_1$ & none \\
\quad\Rmsd & $\bar\tau$ & $\bar d$ & $\bar s$ & $\bar s$ & none \\
\quad\Rko & $t_\mathrm{KO}$ & $q_\mathrm{KO}$ & $p_\mathrm{KO}$ & $p_\mathrm{KO}$ & none \\
\quad\Rkot & $t_\mathrm{KO}$ & $q_\mathrm{KO}$ & $p_\mathrm{KO}$ & $p_\mathrm{KO}$ & tick ratios \\
\addlinespace
\multicolumn{6}{@{}l}{\emph{Raw units}} \\
\quad\Rraw & 1\,s & 100 shares & \$0.01 & \$0.01 & none \\
\quad\Rrawd & 1\,s & 100 shares & \$0.01 & \$0.01 & price, volume rate, volatility, tick \\
\quad\Rrawr & 1\,s & 100 shares & \$0.01 & $\bar s$ & none \\
\quad\Rrawi & 1\,s & 100 shares & \$0.01 & $\bar s$ & window statistics behind the units \\
\addlinespace
\multicolumn{6}{@{}l}{\emph{Standard normalisations}} \\
\quad\Rz & \multicolumn{3}{l}{window mean and s.d.\ of each feature} & $\bar s$ & none \\
\quad\Rratio & $T$ & near-touch size & current spread & $\bar s$ & none \\
\quad\Rdimless & \multicolumn{3}{l}{features without units only} & $\bar s$ & none \\
\bottomrule
\end{tabular}
\end{table}

\paragraph{Learning.}
Because the market path is exogenous, any state on a replayed path can be branched over all five actions, which gives
each action's exact one-step cost and next state on the same path. Fitted Q-iteration~\citep{ernst2005tree} with
gradient-boosted trees, one regressor per decision, learns the advantage of each action over crossing at once, whose
cost the branching gives exactly. All representations share the data, model, hyperparameters and seeds, none tuned
for any of them (\cref{app:learners}). Neural fitted Q-iteration and PPO~\citep{schulman2017ppo} repeat the main
comparisons.

%% file: sec_theory.tex
\section{When should a policy transfer?}
\label{sec:theory}

An asset $a$ generates an exogenous event stream with law $P_a$; because replay is non-intrusive, $P_a$ does not
depend on the agent. A \emph{unit system} $u$ assigns to each asset and episode start a time unit $\tau_u$, a size
unit $q_u$ and a price unit $p_u$, computed from the trailing window only. Rescaling the stream by these units gives a
normalised stream with law $\tilde P^u_a$; a task becomes $(\tilde Q, \tilde T) = (Q/q_u, T/\tau_u)$, and a policy's
normalised cost is $\tilde J = \text{cost}/(Q p_u)$. Four quantities do not rescale, because the exchange or the agent
fixes them in physical units: the tick, the round lot (the standard trading unit), the latency and the fee. Their ratios to the units are the
\emph{dimensionless groups} $g_u(a) = (\text{tick}/p_u, \text{lot}/q_u, \text{latency}/\tau_u, \text{fee}/p_u)$.

\begin{proposition}[Similarity implies invariance]\label{prop:similarity}
If the joint law of the normalised path, the dimensionless groups and the normalised task is the same for assets
$a$ and $b$, then every policy that acts on features expressed in units $u$ has the same expected normalised cost on
$a$ and on $b$. In particular they share their optimal policies.
\end{proposition}

The proof (\cref{app:proofs}) checks that queue accounting, fills and latency commute with rescaling once the four
groups are rescaled consistently. The proposition turns each invariance hypothesis into a prediction about policies,
and says where transfer should break: where the groups differ, above all the tick, which is a quarter of the spread
for some US stocks and the whole spread for others.

\paragraph{How far transfer can fail.}
Real assets are only approximately similar; two results, proved in \cref{app:proofs}, bound what a pooled policy
can lose on a new asset and why some representations must lose.

\begin{proposition}[Zero-shot regret]\label{prop:regret}
Let $S$ be a training mixture of assets, $\Pi$ a policy class, and $\pi_S$ an $\epsilon$-optimal policy in $\Pi$ for the
normalised cost on $S$. For any asset $b$,
$\E_b \tilde J(\pi_S) - \min_{\pi \in \Pi} \E_b \tilde J(\pi) \le \epsilon + 2\, d_\Pi(b, S)$,
where $d_\Pi(b, S) = \sup_{\pi \in \Pi} | \E_b \tilde J(\pi) - \E_S \tilde J(\pi) |$.
Because the paths are exogenous, $d_\Pi(b, S) \le \operatorname{osc}(\tilde J)\, \mathrm{TV}(\tilde P^u_b, \tilde P^u_S)$.
The diagnostic replays a finite family $\Pi_{\text{probe}} \subset \Pi$ of simple rules in units $u$:
$\Dhat_u(b) = \max_{\pi \in \Pi_{\text{probe}}} | \hat\E_b \tilde J(\pi) - \hat\E_S \tilde J(\pi) |$, an estimate of a
lower bound on $d_\Pi$.
\end{proposition}

\begin{proposition}[Aliasing]\label{prop:aliasing}
Let $\phi$ be a representation and $X$ a set of feature values. Suppose that for every $x \in X$, every action is at
least $\Delta$ worse than optimal on all first-decision states of asset $a$ with $\phi = x$, or on all such states of
asset $b$. If the laws of $\phi(s_0)$ under $a$ and under $b$ both put mass at least $\rho\,\mu$ on $X$, for some
probability measure $\mu$ on $X$, then every policy that acts through $\phi$, randomised or not, has regrets on $a$ and
on $b$ that sum to at least $\Delta\rho$.
\end{proposition}

Because the market does not react to the agent, path laws enter the regret bound once instead of compounding
through the agent's state distribution, and $\Dhat_u$ needs no learning on the new asset, only replays of fixed
rules (\cref{sec:res-supply} tests whether it predicts where transfer fails). \Cref{prop:aliasing} predicts that z-scores and depth units lose where assets
differ in queue length per trade size, which they map to the same features.

\paragraph{Why a small transfer gap says little.}
Zero-shot transfer, a pooled policy costing about as much on a new asset as one whose training also saw it, needs
far less than similarity. A comparator that pools many assets gives the new one a small weight $\varepsilon$ in its
data, here about one in two thousand.

\begin{proposition}[A pooled comparator detects little]\label{prop:pooled}
Index the policies of one representation by $\theta$ in a compact set $\Theta \subset \mathbb{R}^p$, and let
$J_x(\theta)$, the expected normalised cost on asset $x$, be twice continuously differentiable on $\Theta$. Let $F$ be
the expected cost on the comparator's training mixture without asset $a$, with a unique minimiser $\theta'$ in the
interior of $\Theta$ and $H = \nabla^2 F(\theta')$ positive definite. Let the seen policy $\theta_\varepsilon$
minimise $(1-\varepsilon) F + \varepsilon J_a$, where $\varepsilon$ is asset $a$'s share of its training data, and let
$\theta_0$ be the zero-shot policy. Then, with $g = \nabla J_a(\theta')$, as $\varepsilon \to 0$,
\[
  \mathrm{TG}_a = J_a(\theta_0) - J_a(\theta_\varepsilon)
  = \big[J_a(\theta_0) - J_a(\theta')\big] + \varepsilon\, g^\top H^{-1} g + o(\varepsilon),
  \qquad 0 \le g^\top H^{-1} g \le \|g\|^2 / \lambda_{\min}(H).
\]
The first term does not involve $a$'s own data; the second, the only part that does, is $O(\varepsilon)$. If moreover
$\nabla J_a$ is $L$-Lipschitz and $\Theta$ contains the ball of radius $\|g\|/L$ about $\theta'$, then
$\varepsilon\, g^\top H^{-1} g \le 2\varepsilon\kappa E_a$, where $\kappa = L/\lambda_{\min}(H)$ and
$E_a = J_a(\theta') - \min_\Theta J_a$ is what the representation's own policies could still save on $a$.
\end{proposition}

The part of a representation's loss that its own policies could save enters the gap with weight at most
$2\varepsilon\kappa$; the part that no policy in the representation can save, such as a loss to aliasing, is not in
the gap at any $\varepsilon$. A transfer gap near zero is therefore necessary but weak evidence: poor units can show
one too. The evidence for units lies in the \emph{contrasts} between representations, each learned and scored the
same way. The proposition assumes a smooth policy class and a comparator that minimises the mixture's cost, as
policy-gradient learners do; for fitted Q-iteration, which minimises a regression loss, it is a heuristic, checked
directly on pendulums: a seen policy for which the test system is half the data recovers
$\PendSweepDiffHalf$ of an excess cost of $\PendSweepExcessRaw$, and no recovery is resolved at the shares a transfer
gap uses (\cref{app:pendulum}).

%% file: sec_design.tex
\section{Evaluation}
\label{sec:design}

\paragraph{Stocks and days.}
The universe is the Nasdaq-traded common stocks priced above \$2 with at least 500 trades a day. It was split once,
before any policy was trained, within strata of listing market, relative tick (the tick over the volatility per
trade) and activity: half of the stocks for training (\UniverseTrain), 15\% for validation and 35\% for testing.
Training uses five 2019 days and validation two later 2019 days. The test sets are held out: three days of 2020--21
(Stage~1; \EpisodesStageOneStocks{} test stocks, including those listed since 2019) and eight days of 2025--26
(Stage~2; \EpisodesStageTwoStocks{} stocks). Each episode draws one of three task families with equal probability, so that no
unit system is favoured by construction: $Q \in \{0.5, 1, 2\}$ mean trade sizes and $T \in \{10, 30, 100\}$ mean times
between trades; the same grid in Kyle--Obizhaeva units; or $Q \in \{100, 300\}$ shares and $T \in \{10, 60, 300\}$
seconds. Every policy runs on the same episodes and the same replayed paths, so every comparison is paired.

\paragraph{Another market and another exchange.}
Coinbase spot pairs differ from stocks in participants, hours and ticks: on US stocks the tick is always one cent, so
a stock's relative tick is tied to its price, while Coinbase sets a tick per pair, and the median relative tick ranges
from 0.004 (BTC) to 2.6 (SHIB). We use 16 pairs quoted in US dollars, from 00:00 to 02:00 UTC on the \CryptoPoolDays{} first-of-month days of
2024--26 and on 1~November 2026, with costs in each pair's volatility per trade (\cref{app:data}). IEX, a second US
stock exchange, publishes its full order-by-order feed daily; we evaluate on twenty days of September and October
2026, the last ten of which had not yet occurred when the policies were frozen, on books whose spread was one tick for at
least half of the trailing 30 minutes (on wider books a crossing order buys through many price levels of IEX's thin
book). The comparators are
the same learners trained on the pairs' 2020--23 days and on five earlier IEX days.

\paragraph{Comparisons and statistics.}
The \emph{seen} policy has the same representation and as many training episodes as the zero-shot one, drawn from the
training stocks and the test stocks' own 2019 days; the \emph{transfer gap} is the zero-shot cost minus the seen cost.
The quantity of interest is a paired cost difference, averaged over three seeds and then over stocks. Intervals are
95\% bootstrap intervals over stocks, stratified by relative tick, or over assets and days for crypto and
IEX~\citep{owen2007pigeonhole}; they hold the trained policies fixed (\cref{sec:res-stocks} adds the variation of
seeds and training stocks). ``Equivalent'' means that the 90\% interval, which amounts to two one-sided tests at the
5\% level~\citep{schuirmann1987tost}, lies within a margin fixed on validation before the test sets were opened: 0.015
spreads per share, a fifth of what learning saves over posting and waiting (resting at the near touch and crossing
what is left at $T$); 0.020 on IEX, the same share of the gain over crossing at once; 0.10 on crypto, not derived
this way.
Analyses added after the test sets were opened are marked \emph{exploratory}.

\paragraph{Implementation.}
The main learner fits one LightGBM regressor per decision (300 rounds, 63 leaves, learning rate 0.05) on 376{,}610
training episodes, with all five actions branched at every decision under a fixed randomised behaviour policy. Its
settings were written with the first version of the learner and never tuned for any representation; refits with more
rounds or leaves help none of them (\cref{app:learners}). The neural learner fits a three-layer MLP per decision, and
PPO trains on about $2\times10^7$ episodes. Every policy is fitted with three seeds; \cref{app:learners} reports
the compute. All data are public (\cref{app:data}), and the code covers every step from raw files to tables and
figures.

\paragraph{Synthetic controls and a second domain.}
A zero-intelligence order book~\citep{daniels2003quantitative, smith2003double}, in which orders arrive, cancel and
execute as independent random streams, has exact symmetries: scaling all
event rates, the order size and the tick value gives an asset whose event stream is the same path, rescaled, and is
exactly similar in microstructure units. Four controls plant known structure: exact similarity (PC1), test scales
outside the training range (PC1x), a varying relative tick (PC2) and varying queue granularity (PC3;
\cref{app:controls}). Outside finance, a family of pendulums that is exactly similar in its natural units, and the
contextual pendulum of CARL~\citep{benjamins2023contextualize}, where similarity is only approximate, ask the same
questions with the same learner (\cref{app:pendulum,app:carl}).

%% file: sec_results.tex
\section{Results}
\label{sec:results}

\paragraph{Controls.}
In the synthetic markets the contrasts between representations mostly behave as theory predicts
(\cref{app:controls}). Under exact similarity raw units cost $\ZiAdvPcOneContrastRZero$ spreads more than \Rms{}
units, and outside the training scales raw and Kyle--Obizhaeva units lose while \Rms{} units transfer
(\cref{sec:res-gap}). Where the relative tick varies, raw units lose $\ZiAdvPcTwoContrastRZero$, and tick context
improves on \Rms{} units for the third of test assets with the most extreme ticks ($\ZiAdvPcTwoContrastExtRThreetick$,
95\% CI $[\ZiAdvPcTwoContrastExtRThreetickLo, \ZiAdvPcTwoContrastExtRThreetickHi]$). One prediction failed: depth
units, which \cref{prop:aliasing} predicts to alias assets of different queue granularity, match \Rms{} units when
it varies ($\ZiAdvPcThreeContrastRThreed$, $[\ZiAdvPcThreeContrastRThreedLo,
\ZiAdvPcThreeContrastRThreedHi]$). The evaluation can tell units apart; the transfer gap alone cannot
(\cref{sec:res-gap}).

\subsection{Q1: Unseen stocks, and six years later}
\label{sec:res-stocks}

\begin{figure}[t]
  \centering
  \safegraphic[width=\textwidth]{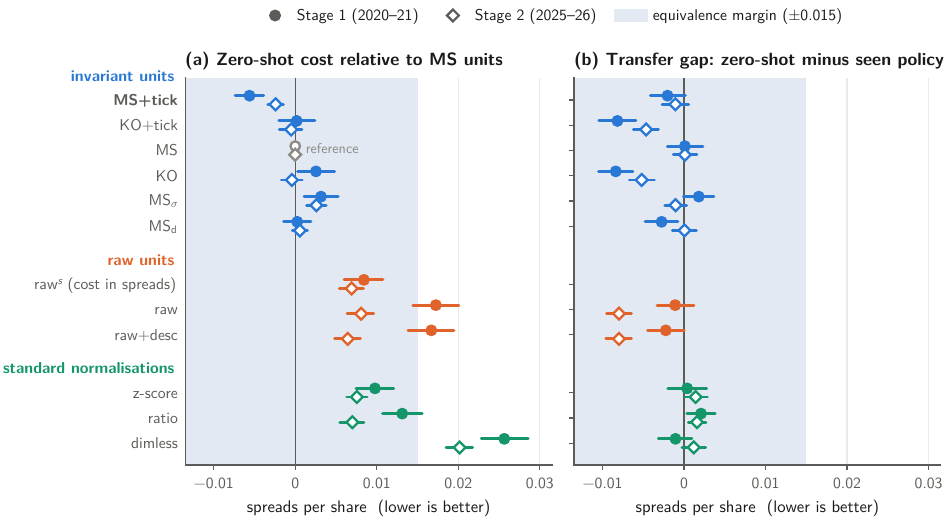}
  \caption{\textbf{Invariant units lead, but every transfer gap is near zero.} (a) Each representation's zero-shot
  cost minus that of \Rms{} units, on held-out stocks in 2020--21 (\EpisodesStageOneStocks{} stocks) and 2025--26
  (\EpisodesStageTwoStocks{} stocks). (b) Its transfer gap: zero-shot cost minus that of the same representation
  trained on as many episodes including the stock's own 2019 days. Three seeds; 95\% stock-bootstrap intervals; band:
  the margin.}
  \label{fig:transfer}
\end{figure}

\textbf{A policy trained on other stocks executes unseen stocks as well as one that saw them, in 2020--21 and in
2025--26.} The zero-shot \Rmst{} policy costs $\StageOneHOneAll$ and $\StageTwoHOneAll$ spreads per share relative to
the seen policy, well inside the margin. Learning pays: in
2020--21 the policy saves $\StageOneGainBpsPost$\,bp of the $\StageOnePostCostBps$\,bp that posting and waiting costs,
and $\StageOneGainBpsImmediate$\,bp relative to crossing at once.

The representations differ by a fraction of that gain (\cref{fig:transfer}a). Raw units ($\StageOneHTwoAll$), z-scores
($\StageOneHThreeAll$) and raw units with the stock's descriptors ($\StageOneHFourAll$) lose about one margin in
2020--21 and half of one in 2025--26, most on large-tick stocks; the dimensionless features alone lose most. The two
invariant unit systems are equivalent, and tick context helps on tasks set in either's units. The transfer gaps, by
contrast, do not separate the units (\cref{fig:transfer}b): raw units' is $\RevStageOneTgSeenRZeroAll$, as
\cref{prop:pooled} leads one to expect (\cref{sec:res-gap}).

\paragraph{Where the tick matters.}
\Cref{prop:similarity} predicts that transfer breaks where the dimensionless groups differ, above all the relative
tick. Given the tick, it does not: the zero-shot \Rmst{} policy matches the seen one in every tercile of the relative
tick ($\StageOneHOneSmall$, $\StageOneHOneMid$ and $\StageOneHOneLarge$ from small to large ticks;
\cref{fig:transfer_stage1}), and tick context improves on \Rms{} units in each ($\StageOneContrastsRThreetickRThreeSmall$
on small-tick stocks, $\StageOneContrastsRThreetickRThreeLarge$ on large-tick ones). Without it the losses concentrate
where the tick binds: raw units lose $\StageOneContrastsRZeroRThreeSmall$ to \Rms{} units on small-tick stocks and
$\StageOneContrastsRZeroRThreeLarge$ on large-tick ones, raw units with descriptors $\StageOneContrastsRZerodescRThreeSmall$
and $\StageOneContrastsRZerodescRThreeLarge$, and z-scores $\StageOneContrastsROneRThreeSmall$ and
$\StageOneContrastsROneRThreeLarge$. Measuring raw units' cost in spreads removes most of their loss on large-tick
stocks ($\StageOneContrastsRZerorRThreeLarge$) but not on small-tick ones ($\StageOneContrastsRZerorRThreeSmall$).
\Cref{app:details} gives the 2025--26 breakdown.

\paragraph{Six years later.}
Between the two periods prices rose, round lots shrank for high-priced stocks and spreads widened
(median $\StageOneMedianSpreadBps$\,bp in 2020--21, $\StageTwoMedianSpreadBps$\,bp in 2025--26). Every contrast keeps
its sign but shrinks, and not because the mix of tick regimes changed: the median stock quotes a spread of
$\StageTwoMedianSpreadTicks$ ticks in both periods, and $\StageTwoOneTickShare$ of stocks are one-tick stocks, against
$\StageOneOneTickShare$. Trained on 2025--26 days of the training stocks (exploratory), \Rmst{}
would gain only $\CtmpStageTwoMstickYears$, about half a margin, and raw units would still lose
$\CtmpStageTwoRawVsMstick$ to it.

\paragraph{Calibrating on the stock does not help.}
Execution is calibrated asset by asset in practice, but the stock's own data add nothing to the zero-shot policy
here (exploratory). On validation, the same learner trained on one stock's five days costs $\PerStockFiveMinus$
spreads more than the pooled zero-shot policy, and a queue rule tuned per stock $\RulesPerStockMinusPooled$ more than
the same rule tuned on the training stocks. Given four features of the stock's 2019 history, \Rmst{} does no better
(zero-shot minus the history-aware policy: $\RFourStageOneGrossMsTickVsHistMsTick$ in 2020--21,
$\RFourStageTwoGrossMsTickVsHistMsTick$ in 2025--26). On validation, a profile of how the stock's orders fill on its
own days changes the cost by $\ProfReal$ ($[\ProfRealSeedLo, \ProfRealSeedHi]$), and per-stock corrections of the
learned advantages by $\RFourValGrossOffsetsMsTickVsMsTick$ (designs in \cref{app:robust}). The policy does use the
stock's own units: given another stock's, it costs $\ValPlaceboRThreetickAll$ more on validation.

\paragraph{Robustness.}
On validation the conclusions hold under both bounds of the fill rule (raw minus \Rms:
$\RobpessimisticContrastsRZeroRThreeAll$ pessimistic, $\RoboptimisticContrastsRZeroRThreeAll$ optimistic,
$\ValContrastsRZeroRThreeAll$ exact), latencies up to 10\,ms and 15- or 60-minute windows (\cref{app:robust}).
Resampling seeds, or stocks and days together, changes no 2020--21 conclusion. Refits on resampled training stocks
vary the contrasts about as much as seeds do, but seeds are averaged and training samples are not: with the test
intervals widened by the whole refit variation, the match with the seen policy holds in both periods, as do z-scores'
loss in both and raw units' loss in 2020--21; the other contrasts hold only under a narrower widening, or not at
all. Over the whole distribution of costs, the zero-shot policy's worst 5\% of episodes cost
$\DistStageOneHOneCvarAbs$ less than the seen policy's, but the units' advantage over raw units holds in the mean
only: at the median raw units are cheaper ($\DistStageOneHTwoMed$), because policies in units wait more, earning the
spread more often and paying more when that fails. A risk-averse objective would rank them the other way.

\subsection{Q2: What the units supply}
\label{sec:res-supply}

\begin{figure}[t]
  \centering
  \safegraphic[width=\textwidth]{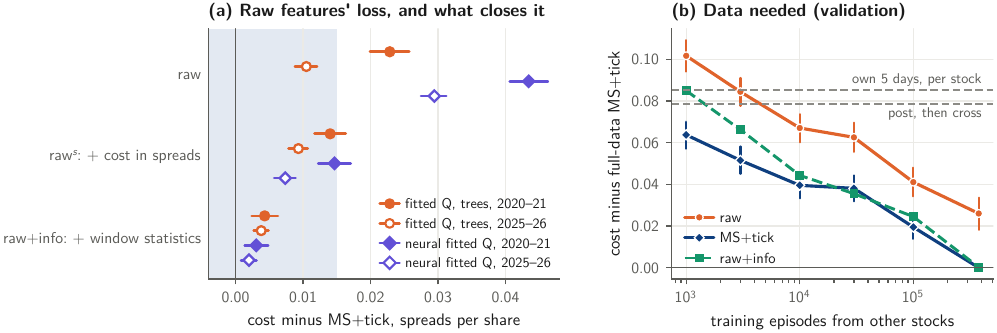}
  \caption{\textbf{The units supply mostly a cost scale and information, and they save data} (exploratory). (a) Raw
  features' cost relative to \Rmst{} on the held-out stocks, as the representation gains a cost in the stock's spread
  (\Rrawr) and then the window statistics the units are built from (\Rrawi), for fitted Q-iteration with trees and
  with a neural network; 95\% stock-bootstrap intervals; the band is the margin. (b) Validation cost against the
  number of training episodes from other stocks, relative to \Rmst{} with the full training set; dashed lines: a
  policy trained on each stock's own five training days, and posting and waiting.}
  \label{fig:supply}
\end{figure}

\textbf{The units supply mostly a common cost scale and information, and they save data.} Measuring the cost in the
stock's spread instead of in cents removes part of raw units' loss, about half on the 2020--21 stocks
(\cref{fig:supply}a), though the share depends on the learner and the training sample (\cref{app:robust}). Adding the window
statistics as plain inputs closes most of the rest, to within about a quarter of a margin of \Rmst{} in both periods
and with both learners. On IEX's wider and thinner books, by contrast, dividing the features helps every learner:
with the window statistics as inputs, raw features still cost $\InfoIexWidePpoInfoVsMstickEpisode$ (PPO) to
$\InfoIexWideTreesInfoVsMstickEpisode$ (trees) more than \Rmst{} with episodes weighted equally, though with stocks
weighted equally only the neural learner's interval excludes zero.

\paragraph{Other learners.}
With neural fitted Q-iteration, which can form a ratio of two inputs, raw units lose more
($\RevStageOneArmNnRawVsMs$ against \Rms), and a per-stock reward scale learned in training, as in
PopArt~\citep{vanhasselt2016popart}, brings raw features within $\RevStageOneArmNnPopVsMsTick$ of \Rmst{}
(\cref{tab:rev-arms}). PPO, an on-policy learner, beats fitted Q-iteration when both use \Rmst{} units ($\PpoMsTickVsTrees$
spreads on validation, $\PpoStageTwoVsTreesMsTick$ in 2025--26), ranks the units the same way (raw units
$\PpoRawVsMsTick$, z-scores $\PpoZVsMsTick$; validation) and is within the margin of its own seen policy
($\PpoTgMsTick$, 90\% CI $[\PpoTgMsTickLo, \PpoTgMsTickHi]$). It halves raw units' loss on validation, not in
2025--26 ($\PpoStageTwoRawVsMsTick$; trees $\TreesStageTwoRawVsMsTick$).

\paragraph{Learned alternatives.}
Three ways of learning what the units supply, tried with the neural learner on validation days with the cost in
spreads, do no better. Invariance penalties, with the nine cells of price level by trade rate as environments,
leave raw features' cost unchanged or raise it (IRMv1~\citep{arjovsky2019irm} $\InvIrmOneRawrVsPlain$,
V-REx~\citep{krueger2021rex} $\InvVrexOneRawrVsPlain$, GroupDRO~\citep{sagawa2020groupdro} $\InvGdroTenRawrVsPlain$). A
context encoder~\citep{rakelly2019efficient} that reads the trailing 30 minutes as per-minute trade counts, volumes,
spreads, depths and returns does barely better than raw features ($\CtxVsRawr$) and worse than the window statistics
as inputs ($\CtxVsInfo$); on IEX's wider books it recovers more than half of raw features' loss but stays
$\IexOodCtxVsMstick$ behind \Rmst. Domain randomisation~\citep{tobin2017domain}, stretching event times by 1/4 to 4 in the replay and
rescaling prices and sizes, raises raw features' cost ($\TdrRawrVsPlain$). Each is one architecture, untuned beyond
two strengths.

\paragraph{Data, money and extrapolation.}
What the units do save is data (\cref{fig:supply}b): raw units need about ten times the episodes to reach the
cost of \Rmst. In money the differences are small: raw units cost $\RevStageOneEconRawVsMsBps$\,bp more than \Rms{}
before fees and nothing after Nasdaq's fees, because they rest more orders, and Nasdaq pays resting orders that fill
a rebate and charges crossing ones a fee (\cref{tab:rev-econ}). In
2025--26 raw units are cheaper after fees ($\RevStageTwoEconRawVsMsBpsNet$\,bp, $[\RevStageTwoEconRawVsMsBpsNetLo,
\RevStageTwoEconRawVsMsBpsNetHi]$), though dearer before them, the cost the policies were trained on.
Nor do the units extrapolate (exploratory). Trained only on stocks in the middle half of each scale's range and
scored on 2020--21 stocks outside it, no representation shows a gap across a \XtrapDescActivityOutsideRatio-fold range
of activity, but \Rmst{} has gaps of $\RevStageOneXabsTradesizeGapRThreetickAbove$ beyond the training trade sizes
and $\RevStageOneXabsSpreadGapRThreetickOutside$ outside the spread range; raw units, worse everywhere, degrade less.
The RL-free diagnostic of \cref{prop:regret}, replaying
22 simple rules on each stock's 2019 days, does not predict where transfer fails (rank correlation
$\XtrapSpreadRThreetickDhatrho$ with the per-stock loss outside the spread range, on validation stocks;
\cref{app:robust}).

\subsection{Q3: Another market and another exchange}
\label{sec:res-markets}

\begin{figure}[t]
  \centering
  \safegraphic[width=\textwidth]{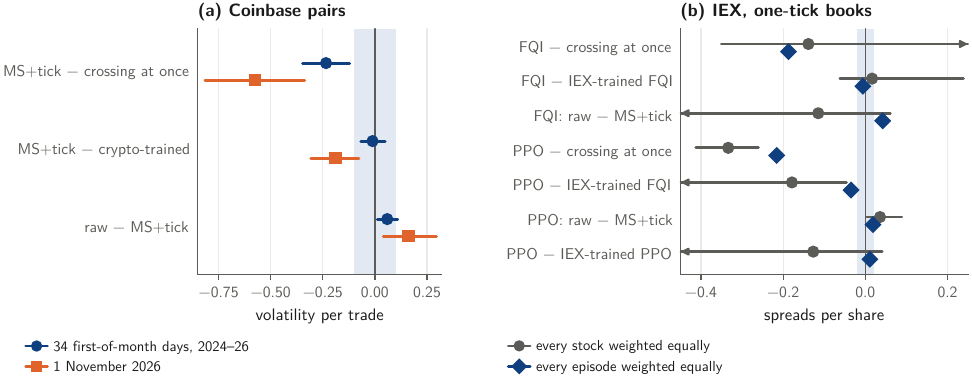}
  \caption{\textbf{The stock-trained policies transfer to crypto and to a second exchange.} Paired cost differences;
  bands: the margins; arrows: intervals beyond the axis. (a) Coinbase pairs, in volatility per trade, 90\% intervals.
  (b) The ten IEX days after the policies were frozen, books at one tick, in spreads per share, 95\% intervals, with
  every stock or every episode weighted equally. FQI: fitted Q-iteration.}
  \label{fig:markets}
\end{figure}

\paragraph{Crypto.}
In each pair's volatility per trade, the stock-trained \Rmst{} policy costs $\CryptoPoolCOneAll$ relative to
crossing at once over the \CryptoPoolDays{} days and is within the margin of a policy trained on the pairs
($\CryptoPoolCTwoAll$, 90\% CI $[\CryptoPoolCTwoAllLo, \CryptoPoolCTwoAllHi]$); its units beat raw units
($\CryptoPoolCThreeAll$; \cref{fig:markets}a). On 1~November 2026 it is cheaper than the crypto-trained policy
($\CryptoProspNovCTwoAll$). On the twelve quarterly days alone, the comparison with the crypto-trained policy is too
noisy to decide ($\CryptoTestCTwoAll$, $[\CryptoTestCTwoAllLo, \CryptoTestCTwoAllHi]$), because costs vary from day to
day; on the 22 other first-of-month days it is $\CryptoRepCTwoAll$ ($[\CryptoRepCTwoAllLo, \CryptoRepCTwoAllHi]$). A
reference trained on half of the 2024--26 days is no stronger: the stock-trained policy costs
$\CryptoXZeroVsRef$ relative to it. Crypto costs are in volatility per trade, chosen on development days, because in
spreads a few pairs whose one-tick spread is far below their volatility would dominate every average.

\begin{figure}[t]
  \centering
  \safegraphic[width=\textwidth]{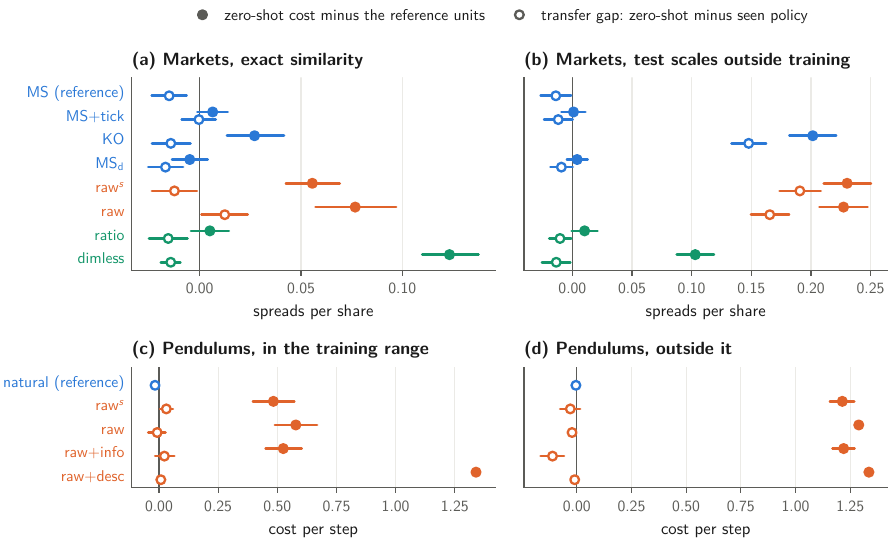}
  \caption{\textbf{Wrong units cost more without a larger transfer gap, unless the comparator can tell the new
  assets apart.} For each representation, its zero-shot cost minus that of the invariant units (filled) and its
  transfer gap (open). (a, b) Synthetic order-book markets, exactly similar in \Rms{} units, with test assets inside
  and outside the training scales; spreads per share. (c, d) Pendulums, exactly similar in natural units, with test
  systems inside and outside the training range of time scales; cost per step. 95\% bootstrap intervals over test
  assets or systems.}
  \label{fig:lesson}
\end{figure}

\paragraph{IEX.}
On the ten IEX days that followed the freezing of the policies, zero-shot fitted Q-iteration and zero-shot PPO are
within the margin of the same learners trained on IEX when every episode counts equally ($\IexTestEpisodeITwoAll$ and
$\IexTestLmEpisodePFourAll$; \cref{fig:markets}b), and PPO beats crossing and raw units. With every stock counting
equally the intervals are wide: the stock-weighted mean has an effective sample size of \NeffIexTestShare{} of the
\AnalysedIexTest{} episodes (\NeffStageOneShare{} on Nasdaq), and one stock's only episode, whose forced cross sold
far below the market, carries \IexDiagWeight{} times an average episode's weight; order protection would have
prevented that trade (\cref{app:glossary}). Without it, fitted
Q-iteration's stock-weighted comparison with its IEX-trained counterpart takes the sign of the episode-weighted one
($\IexNoNhpITwo$, $[\IexNoNhpITwoLo, \IexNoNhpITwoHi]$). PPO's gap
to the IEX-trained PPO is small, but its 95\% interval excludes zero ($[\IexRePFourEpisodeLoNinetyfive,
\IexRePFourEpisodeHiNinetyfive]$)
and would not fit within the Nasdaq margin of 0.015. It is mostly what IEX's own 2025--26 data add: PPO trained on
Nasdaq's one-tick books alone recovers about a quarter of it, and fine-tuning on IEX's books all of it (fine-tuned
minus zero-shot $\FtTestFtVsZeroEpisode$; exploratory). On wider IEX books raw
units, z-scores and raw units with a learned reward scale all lose to \Rmst{} (by $\IexWideWFiveAll$ to
$\IexWideWFourAll$ with stocks weighted equally, $\IexTestWideEpisodeWFiveAll$ to $\IexTestWideEpisodeWFourAll$ with
episodes), where the single-venue replay is least realistic (\cref{app:iex}).

\subsection{Q4: Does a small transfer gap certify the units?}
\label{sec:res-gap}

\textbf{Wrong units cost more without a larger transfer gap, unless the comparator can tell the new assets apart.}
Under exact similarity in the synthetic markets, raw units cost $\ZiAdvPcOneContrastRZero$ spreads more than \Rms{}
units while their transfer gap is $\ZiAdvPcOneTgRZero$ (\cref{fig:lesson}a); on the Nasdaq stocks, raw units' gap is
$\RevStageOneTgSeenRZeroAll$ against a loss of $\StageOneHTwoAll$ (\cref{fig:transfer}). This is the small share at
work (\cref{prop:pooled}): a pooling comparator barely adapts to any one asset. Outside the training
scales the gap does flag raw and Kyle--Obizhaeva units ($\ZiAdvPcOnexTgRZero$ and $\ZiAdvPcOnexTgRTwo$;
\cref{fig:lesson}b), because a comparator that saw the test assets can tell them apart.

Pendulums show the same outside finance. A point mass on a rod, with mass, length and gravity drawn over a range, is
exactly similar in its natural units (time $\sqrt{l/g}$, torque $mgl$), and the task is to swing it up and balance it
with a limited torque (\cref{app:pendulum}). With the same learner, raw features cost $\PendInRangeRawVsNat$ more per
step than natural units in range, with a transfer gap of $\PendInRangeTgRaw$, and $\PendOutRangeRawVsNat$ more out of
range, with a gap of $\PendOutRangeTgRaw$ (\cref{fig:lesson}c, d). Out of range the gap misses the failure for a
different reason: in raw features the states that decide a swing-up, slow motion near the top and the bottom, look
alike for every pendulum, so the seen policy fails in the same way (aliasing; \cref{prop:aliasing}). On the
contextual pendulum of CARL~\citep{benjamins2023contextualize}, where similarity is only approximate, the neural
learner's gap for raw features is a quarter of their excess cost ($\CarlNnInTgRaw$ against $\CarlNnInRawVsNatc$;
\cref{app:carl}).

%% file: sec_discussion.tex
\section{Conclusion}
\label{sec:discussion}

We tested microstructure invariance on decisions: policies trained once on 2019 Nasdaq stocks were evaluated
unchanged on held-out stocks, later years, crypto pairs and a second exchange.

\paragraph{What a policy needs to know about the asset it trades.}
Within the scales its training covered: the asset's recent trade rate, trade size and spread, and its tick, on a
common cost scale. Given these, a policy trained on other stocks matches one that also saw the stock. Whether the
quantities enter as units or as inputs matters little on familiar books; the units save data and help on thinner
books. Nothing we tried carries a policy into unseen spread regimes, so training data must span the spreads
a policy will meet.

\paragraph{Lessons for evaluating transfer.}
Two observations may carry over wherever one policy serves tasks of different scales. First, a near-zero gap to a
comparator that pools the test task with many others is weak evidence that a representation is right, because such a
comparator adapts to any one task only in proportion to its share of the data (\cref{prop:pooled,sec:res-gap}).
Evaluations that rest a representation, a context encoder or a normalisation on such a gap should also compare
alternatives trained and scored alike, or give the test task a large share of the comparator's data. Second, an average weighting
every task equally can hinge on one heavy-tailed episode, as on IEX; reporting task- and
episode-weighted averages exposes this.

\paragraph{Limitations and future work.}
The replay is one venue's displayed book, non-intrusive and decided at ten fixed times: historical orders do not react
to the agent, and there is no routing, order protection, or hidden or midpoint liquidity. A cross walks only the
venue's displayed depth, and the order exceeds the displayed size at the opposite best in \ReplayBeyondNasdaq{} of
Nasdaq training episodes, so crossing costs more than it would for a routed order: costs relative to crossing at once
are overstated, and so may be contrasts between policies that cross different amounts. Costs are gross of fees unless stated,
the learned alternatives to the units were tried on validation days only, and the data are public sample days and
few crypto pairs. Make-or-take is only the atom of execution; splitting a parent order, where impact matters, is
the next step. The US half-penny tick, postponed past our data~\citep{sec2025exemptive}, will halve the tick of the
most tick-constrained stocks at once: a direct test of tick context.

%% file: shared/impact.tex
Better execution lowers trading costs, here by a few basis points per order; the methods enable nothing existing
algorithms do not, and the data are public, with no personal data. The replay cannot show how markets would change
if many traders used one rule across many instruments, which transfer makes easier.

%% file: appendix.tex
\appendix
\counterwithin{figure}{section}
\counterwithin{table}{section}
\input{app_glossary}
\input{app_data}

\input{app_learners}

\input{app_proofs}
\input{app_controls}
\input{app_results}
\input{app_pendulum}
\input{app_carl}

%% file: app_glossary.tex
\section{Glossary, representations and policies}\label{app:glossary}

\paragraph{Market terms.} \emph{Limit order book}: the resting buy orders (bids) and sell orders (asks) of an asset at
each price. \emph{Touch}: the best bid and the best ask; the \emph{near touch} is the best price on the agent's own
side, the \emph{far touch} the best price on the other side. \emph{Spread}: the best ask minus the best bid.
\emph{Tick}: the step of the price grid, one cent for US stocks above \$1; \emph{relative tick}: the tick over the
volatility per trade, so that large-tick stocks usually quote a one-tick spread. \emph{Queue position}: the shares
resting ahead of an order at its price, which fill first. \emph{Crossing} (taking): trading at once against resting
orders on the other side; \emph{posting} (making): resting an order and waiting. \emph{Rebate and fee}: Nasdaq pays
resting orders that fill a rebate and charges orders that cross a fee. \emph{Displayed book}: the visible orders;
hidden and midpoint orders are not shown. \emph{Round lot}: the standard trading unit, 100 shares for most stocks and
fewer for high-priced ones since late 2025. \emph{Trade-through}: a trade at a price worse than a resting order's,
which the resting order would have filled first. \emph{Order protection}: the US rule that bars an exchange from
trading at a worse price than another exchange displays. \emph{Basis point} (bp): 0.01\% of the price.

\paragraph{Unit definitions.} \Cref{tab:units} gives the units of each representation. Kyle and Obizhaeva's units are $t_\mathrm{KO} = c_t x^{-2/3}$,
$q_\mathrm{KO} = c_q V x^{-2/3}$ and $p_\mathrm{KO} = c_p P\sigma x^{-1/3}$. Here $P$ is the price, $V$ the daily
volume and $\sigma$ the daily volatility, each scaled from the window; $x = W/W^*$ is the trading activity
$W = PV\sigma$ relative to the benchmark $W^* = 0.02 \cdot 40 \cdot 10^6$; and the constants $c$ make the median
\Rko{} units equal the median \Rms{} units on the training stocks. The window statistics that \Rrawi{} adds are those
the \Rms, \Rmss{} and \Rko{} units are built from: the mean time between trades, trade size and spread; the price,
volume rate, one-minute and per-trade volatility; and the tick. \Rrawi{} was added after the main evaluation.
\Rratio{} takes its flow and trade counts over the last decision interval, $T/K$, and \Rdimless{} keeps the touch
imbalance, the time and shares left and the order's status.

\paragraph{Policies.} \emph{Zero-shot}: trained on the training stocks only. \emph{Seen}: the same representation and
as many training episodes, drawn from the training stocks and the test stocks' 2019 days (Nasdaq), or trained on the
test venue (crypto, IEX). Learners: fitted Q-iteration with trees (the main learner), neural fitted Q-iteration and
PPO (\cref{tab:learners}). Baselines: crossing at once, posting and waiting (posting once at the near touch and
crossing what is left at the end), and re-pegging to the touch.

%% file: app_data.tex
\section{Data pipeline and replay validation}\label{app:data}

\paragraph{From ITCH messages to event tables.}
Each sample day is a binary stream of Nasdaq TotalView-ITCH 5.0 messages for every symbol traded on Nasdaq
($\DataSymbolsMin$ to $\DataSymbolsMax$ per day). We index the stream once and group messages by stock. A second
compiled pass then writes one event table per stock. Its rows are additions, executions, partial cancels, deletes and
replaces (as a delete followed by an add at the back of the queue), hidden prints, auction crosses and trading-state
changes. Orders resting at 09:25 are written as one snapshot in time priority. The $\DataDays$ days processed so far hold
\DataMessages{} messages and \DataRows{} event rows; of the 19 public days, the half day of 2025-11-28 is not used.

\paragraph{Ticks, fees and round lots.} No displayed price above one dollar is off the penny grid in the
\OffTickItchRows{} Nasdaq rows (\OffTickItchDays{} days) or the \OffTickIexRows{} IEX rows (\OffTickIexDays{} days); the
half-penny increment (Rule 612) and the lower access-fee cap (Rule 610(c)) adopted in 2024 were postponed past our last
day~\citep{sec2025exemptive}, so the 30-mil cap and our fee schedule hold. The new round lots took effect in November
2025; the eligibility rule's 100-share round lot drops \RoundLotStageTwoShare{} of the 2019 test-split stocks'
otherwise eligible 2025--26 days (\RoundLotStageTwoStocks{} stocks, median price \$\RoundLotStageTwoMedianPrice) and
\RoundLotIexTestShare{} of all stocks' IEX test days (\RoundLotIexTestStocks{} stocks; round lots from the 12 June 2026
Nasdaq directory).

\paragraph{Book consistency.}
Replaying each table through a price-time-priority book gives, on every day,
\begin{itemize}
  \item $\DataUnresolved$ executions or cancels of orders that are not resting;
  \item no delete that removes a different number of shares than were left;
  \item $\DataCrossedRows$ rows in continuous trading after which Nasdaq's own book was locked or crossed, across all
  days;
  \item executions at the displayed price that hit the oldest order at their level in $\DataFifoMin$ to
  $\DataFifoMax$ of cases.
\end{itemize}
Executions reported at a price other than the displayed one come from discretionary orders. They are counted
separately and never fill a virtual order. All $\DataDaysPassed$ days pass every gate.

\paragraph{Tracker against an independent implementation.}
A toy matching engine with price-time priority produces random message streams with sweeps, partial cancels, replaces,
executions away from the displayed price and hidden prints. The streams are encoded as ITCH and processed by the same
pipeline. On each stream, random virtual orders (at the touch, inside the spread or behind it) are tracked twice:
\begin{itemize}
  \item by the fast tracker, which works on event tables and creation rows;
  \item by an independent tracker that inserts the order into explicit FIFO queues rebuilt from the raw messages.
\end{itemize}
They agree fill for fill (time, size and price) on every order of 30 streams.

\paragraph{Virtual twins of real orders.}
For $\FillOrders$ real orders that joined the touch during continuous trading ($\FillStockDays$ stock-days), we insert a
virtual order with the same side, price and size at the real order's own queue position. The twin's first fill falls
exactly on the real order's first execution in $\FillTwinMatch$ of the cases where the real order was executed (only
$\FillExecutedShare$ of touch orders are). Every exception is a designed rule: an execution away from the displayed price,
or a hidden print at the order's price once nothing is ahead. A second twin placed right behind the real order
first fills before it in $\FillTwinBehindFirst$ of cases.

\paragraph{Coinbase order-by-order data.}
Coinbase's spot exchange publishes every order event of its books with a sequence number that runs across all message
types. We use the free first-of-month files of Tardis.dev, which record the exchange's websocket feed together with a
full snapshot of the book at every (re)connection. Each pair-day is rebuilt as follows.
\begin{itemize}
  \item The book is seeded from the first snapshot, and messages are applied in sequence order from there.
  \item An open order becomes an addition, a match an execution of the resting maker, and a done the removal of
  what is left.
  \item A size cut at the same price keeps priority. Any other modification (a new price or a larger size) moves the
  order to the back of its new level.
  \item An order modified through the market arrives as its own taker matches first, then a change that rests the
  remainder. The old order therefore leaves its level at the first of these matches.
  \item Any gap in sequence numbers is a websocket reconnect. Even one lost cancel leaves a stale order, so the outage
  is marked (no episode may overlap it) and the book is reconciled with the reconnect's snapshot.
  \item Prices and sizes are parsed exactly and stored in ticks and lots of each pair-day: the greatest common
  divisors of all its prices and sizes. In units of $10^{-8}$ coins, large-supply coins overflow 64-bit integers.
\end{itemize}
Over $\CryptoQaPairDays$ pair-days on $\CryptoQaDays$ days ($\CryptoQaPairs$ pairs, \CryptoQaMessages{} messages),
the replay finds no unresolved reference, no negative size and $\CryptoQaCrossed$ locked or crossed rows. The
$\CryptoQaExecutions$ executions include $\CryptoQaFifoBad$ that did not hit the oldest order at their level.
$\CryptoQaReseeds$ reconnects were bridged, for $\CryptoQaOutageMinutes$ minutes of outage in total. A handful of done
messages report a remaining size that disagrees with the order's own fills, although the sequence is complete. These
are inconsistencies in the feed; the replay removes whatever the fills left.

\input{shared/detail/replay_limits}

%% file: shared/detail/replay_limits.tex
\paragraph{What the replay does not represent.}
The queue bookkeeping is exact; the market around it is simplified. The replay is one venue's book in isolation:
there is no national best quote, order protection or routing, and the mid and spread are the venue's own. A cross
takes displayed depth only, walking up to 50 non-empty price levels at any distance from the touch; shares beyond them are charged at the last level plus five
ticks, and if the opposite side shows nothing, at the near touch plus six ticks. It never meets hidden or midpoint
liquidity, which a routed marketable order often would. The order often exceeds the displayed size at the opposite
best on one venue (in \ReplayBeyondNasdaq{} of Nasdaq training episodes, median ratio \ReplayBeyondNasdaqMedian; in
\ReplayBeyondIexOnetick{} on IEX books at one tick and \ReplayBeyondIexAll{} on all IEX books), so crossing is
dearer here than for a routed order, and more so for policies that cross more: costs relative to crossing at once are
overstated, and so may be contrasts between policies that cross different amounts. A cross runs beyond the visible
book, and the shortfall rule fires, in \ReplayShortNasdaqVal{} of \ReplayShortNasdaqValEpisodes{} Nasdaq validation
episodes, but in \IexReIOneShortfallEpisodes{} of the \AnalysedIexTest{} one-tick IEX test episodes and in
\IexReWOneShortfallEpisodes{} of the \AnalysedIexTestWide{} wide-book ones (episodes in which any policy of the
fitted-Q evaluation fell short; \IexRePOneShortfallEpisodes{} one-tick episodes in the PPO evaluation). A walk can also
reach a single resting order far from the market, which order protection would route around. For the fifth of
validation episodes whose order is at most a quarter of the mean touch depth, the main contrasts keep their signs,
with wide intervals (\cref{app:results}). Passive orders are displayed limit orders at the touch or one tick inside; there are no pegged, hidden or midpoint orders. Decisions come at ten fixed times, not on events, and
historical orders do not react to the agent. Where these limits bind, on IEX's thin books, we say so.

%% file: app_learners.tex
\section{Learners and compute}\label{app:learners}
\Cref{tab:learners} lists every setting of the three learners. None was tuned for a representation. The tree
settings were written with the first version of the learner and never changed; a pilot on 300 training stocks ran them
unchanged with raw, Kyle--Obizhaeva, microstructure and microstructure-plus-tick units to choose between direct and
advantage targets, and advantage targets were better for all four. The neural settings were fixed when that learner
was added, and the PPO settings after one smoke run in microstructure-plus-tick units.
A capacity and tuning check closes this appendix (exploratory). The two learners differ in more than the algorithm:
fitted Q-iteration learns from a fixed randomised behaviour policy on 376{,}610 branched episodes, PPO from its own
state distribution over about $2\times10^7$ episodes, with a robust value loss; the target of fitted Q-iteration takes
a minimum over estimated advantages, which is biased downwards; and PPO's settings were fixed after a smoke run in
\Rmst{} units, while the tree settings were never tuned. Comparisons between the learners are comparisons of these two
configurations.

\begin{table}[h]
\centering
\small
\caption{The learners. Episodes: the training stocks' five 2019 days (4{,}431 stock-days, 100 start times each, before
the horizon and closing-time rules). Times: wall clock on a shared machine.}
\label{tab:learners}
\setlength{\tabcolsep}{4pt}
\begin{tabular}{@{}>{\raggedright\arraybackslash}p{0.09\textwidth}>{\raggedright\arraybackslash}p{0.29\textwidth}%
>{\raggedright\arraybackslash}p{0.27\textwidth}>{\raggedright\arraybackslash}p{0.29\textwidth}@{}}
\toprule
 & fitted Q-iteration, trees (main) & fitted Q-iteration, neural (exploratory) & PPO \\
\midrule
model & LightGBM regression, one per decision & MLP, 3 hidden layers of 256 SiLU units, one per decision & policy and
value MLPs, 2 hidden layers of 128 SiLU \\
settings & 300 rounds, 63 leaves, learning rate 0.05, $\ge 200$ samples per leaf, feature fraction 0.9, bagging 0.8,
L2 penalty 1, deterministic & AdamW, learning rate $10^{-3}$, weight decay $10^{-5}$, batch 4{,}096, up to 40 epochs,
early stopping (patience 4) on 5\% of episodes held out & clip 0.2, GAE $\lambda = 0.95$, no discounting, 4 epochs of
minibatches of 16{,}384, Adam $3\cdot10^{-4}$, entropy bonus 0.01 to 0.001, Huber value loss, gradient norm clipped at
1, advantages standardised \\
inputs & the representation's features & $\operatorname{sign}(x)\log(1+|x|)$, standardised & as the neural learner,
plus the decision index \\
target & advantage over crossing now, squared error & the same, scaled to unit RMS per decision & the episode cost
(policy gradient) \\
data & 376{,}610 episodes, all five actions branched at each decision; behaviour policy \textsc{hold} 40\%,
\textsc{join} 25\%, \textsc{improve} 10\%, \textsc{cancel} 10\%, \textsc{cross} 15\% & the same & 150 iterations,
each on 35\% of the stock-days, about 130{,}000 episodes per iteration \\
seeds & 3 (8 in some analyses) & 5 & 3 \\
compute & about 2.5 min per fit, 16 CPU threads & about 1 min per fit, one Quadro RTX 6000 & 12 to 40 min per run,
12--16 CPU workers and one GPU \\
\bottomrule
\end{tabular}
\end{table}

\paragraph{Total compute.} A tree fit takes about 2.5 minutes on 16 CPU threads. Over the whole project, including
every exploratory arm, the run logs
record about 80 hours of wall-clock time for 1{,}564 fitted-Q fits (16 to 32 CPU threads each), 30 hours for 60 PPO
runs, 10 hours for 169 evaluation runs (up to 32 worker processes) and under three hours for building, checking and
sampling the replay tables, on one shared 64-thread machine with one GPU in use; downloads are not counted.

\paragraph{Capacity and tuning check (exploratory, validation).}
Because the settings were not tuned, we refitted every tree arm with more capacity (\cref{tab:capacity}) and PPO with
its value targets on a common scale. More capacity does not help raw units (1{,}000 rounds instead of 300:
$\InfoTuneRoundsRawVsBase$, $[\InfoTuneRoundsRawVsBaseLo, \InfoTuneRoundsRawVsBaseHi]$; 255 leaves instead of 63:
$\InfoTuneLeavesRawVsBase$; both: $\InfoTuneBothRawVsBase$) and makes \Rmst{} worse ($\InfoTuneRoundsMstickVsBase$,
$\InfoTuneLeavesMstickVsBase$ and $\InfoTuneBothMstickVsBase$); with 1{,}000 rounds that loss is about two standard
deviations across seeds, so reading it as overfitting needs some caution. The default setting is the best of the
four, or within noise of it, for every arm. With each arm at its best capacity (255 leaves for raw units, the
default setting for \Rmst), raw units lose by $\InfoTuneBestRawVsMstick$ ($[\InfoTuneBestRawVsMstickLo,
\InfoTuneBestRawVsMstickHi]$), about the same as at the default settings ($\InfoTreesRawVsMstick$); choosing each
arm's best on the days that score it flatters both arms a little. At matched larger capacity every arm whose cost is in
spreads is worse than at its default setting, and raw units in cents only with both changes
($\InfoTuneBothRawVsBase$), so the contrasts move both ways: with 1{,}000 rounds \Rrawi{} falls behind \Rmst{}
($\InfoTuneRoundsInfoVsMstick$, against $\InfoTreesInfoVsMstick$ at the default), while with 255 leaves, or both
changes, raw units and \Rrawi{} come within noise of an \Rmst{} that has itself lost $\InfoTuneLeavesMstickVsBase$
and $\InfoTuneBothMstickVsBase$. Contrasts between overfitted learners say little about the units. Dividing PPO's value targets by the spread of its first returns,
so that the critic's loss is in the same regime whatever the reward unit, helps \Rmst{} a little
($\InfoPpoVsMidMstickVsBase$ at the default learning rate) and raw units not measurably ($\InfoPpoVsMidRawVsBase$;
at learning rates of $10^{-4}$ and $10^{-3}$, $\InfoPpoVsLowRawVsBase$ and $\InfoPpoVsHighRawVsBase$); under it raw units
still lose by $\InfoPpoVsMidRawVsMstick$, $\InfoPpoVsLowRawVsMstick$ and $\InfoPpoVsHighRawVsMstick$ at the three
learning rates. With the window statistics as inputs and the cost in spreads, PPO matches \Rmst{} on validation
($\InfoPpoInfoVsMstick$, $[\InfoPpoInfoVsMstickLo, \InfoPpoInfoVsMstickHi]$). PPO's entropy schedule was not varied.

\begin{table}[!htbp]
\centering
\small
\caption{Capacity of the tree learner (validation, exploratory). Top: each arm's cost relative to its default
setting; bottom: each arm minus \Rmst{} at the same setting. Spreads per share, stocks weighted equally, 95\% intervals
(stock bootstrap stratified by relative-tick tercile). One seed per larger setting, three for \Rmst{} with 1{,}000
rounds; the default column averages its three seeds, so seed variation (a standard deviation of about
0.004 for a validation contrast) is not in the intervals of the other columns.}
\label{tab:capacity}
\IfFileExists{tab_capacity.tex}{\input{tab_capacity.tex}}{(pending)}
\end{table}

%% file: tab_capacity.tex
\begin{tabular}{@{}lllll@{}}
\toprule
 & default & 1{,}000 rounds & 255 leaves & both \\
\midrule
\multicolumn{5}{@{}l}{\emph{cost relative to the arm's default setting}} \\
\Rmst & 0 & $+0.009$ {\scriptsize[$+0.003$, $+0.016$]} & $+0.019$ {\scriptsize[$+0.010$, $+0.027$]} & $+0.049$ {\scriptsize[$+0.038$, $+0.059$]} \\
\Rraw & 0 & $+0.001$ {\scriptsize[$-0.004$, $+0.006$]} & $-0.002$ {\scriptsize[$-0.008$, $+0.003$]} & $+0.026$ {\scriptsize[$+0.015$, $+0.037$]} \\
\Rrawr & 0 & $+0.018$ {\scriptsize[$+0.010$, $+0.025$]} & $+0.017$ {\scriptsize[$+0.009$, $+0.025$]} & $+0.052$ {\scriptsize[$+0.041$, $+0.064$]} \\
\Rrawi & 0 & $+0.019$ {\scriptsize[$+0.012$, $+0.027$]} & $+0.018$ {\scriptsize[$+0.011$, $+0.025$]} & $+0.048$ {\scriptsize[$+0.038$, $+0.058$]} \\
\midrule
\multicolumn{5}{@{}l}{\emph{minus \Rmst{} at the same setting}} \\
\Rraw & $+0.026$ {\scriptsize[$+0.018$, $+0.034$]} & $+0.017$ {\scriptsize[$+0.007$, $+0.028$]} & $+0.005$ {\scriptsize[$-0.006$, $+0.016$]} & $+0.003$ {\scriptsize[$-0.008$, $+0.015$]} \\
\Rrawr & $+0.012$ {\scriptsize[$+0.005$, $+0.018$]} & $+0.020$ {\scriptsize[$+0.012$, $+0.028$]} & $+0.010$ {\scriptsize[$+0.001$, $+0.019$]} & $+0.015$ {\scriptsize[$+0.005$, $+0.025$]} \\
\Rrawi & $+0.002$ {\scriptsize[$-0.004$, $+0.007$]} & $+0.012$ {\scriptsize[$+0.004$, $+0.020$]} & $+0.001$ {\scriptsize[$-0.009$, $+0.010$]} & $+0.001$ {\scriptsize[$-0.009$, $+0.010$]} \\
\bottomrule
\end{tabular}

%% file: app_proofs.tex
\section{Proofs}\label{app:proofs}

Throughout, an episode's market path $\omega$ is exogenous. Given $\omega$, the task $(Q, T, s)$, the physical
constants (tick, lot, latency, fee) and the agent's actions, the replay is a deterministic map $F$ to fills,
observations and costs (\cref{sec:setup}). A policy may randomise; its randomness is independent of $\omega$.

\paragraph{Proof of \cref{prop:similarity}.}
Let $S$ be the map that rescales a path by the units $(\tau, q, p)$ of system $u$. It sends times $t \mapsto (t - t_0)/\tau$,
sizes $x \mapsto x / q$ and prices $y \mapsto (y - m_0)/p$, and applies the same map to the task, $(Q, T) \mapsto (\tilde Q, \tilde T)$,
and to the constants, which become the groups $g_u$. Every operation of $F$ is invariant under $S$:
\begin{itemize}
  \item placing an order at the touch or one tick inside, and deciding whether it is marketable, compares prices on the
  tick grid;
  \item the shares ahead and the fills add and compare share counts;
  \item latency and the decision schedule add and compare times;
  \item a cost is a sum of (price $-$ arrival mid) $\times$ shares, divided by $Q p$;
  \item features in units $u$ are, by definition, functions of the rescaled quantities.
\end{itemize}
Hence $F(S\omega, a, S(\text{task}), g_u) = S\,F(\omega, a, \text{task}, \text{constants})$ for every action sequence $a$.
A policy acting on features in units $u$ therefore produces, given the groups and the normalised task, a law of
observations, actions and normalised costs that depends on the asset only through the conditional law of $S\omega$.
Integrating over the joint law of $(S\omega, g_u, \text{normalised task})$, which is the same for $a$ and $b$, gives
the same joint law on both assets. Its expected normalised cost is then the same on $a$ and $b$, and so is the set of
minimisers. $\square$

\paragraph{Proof of \cref{prop:regret}.}
Write $J_b(\pi) = \E_b \tilde J(\pi)$ and $d = d_\Pi(b, S)$, and let $\pi^\star_b$ minimise $J_b$ over $\Pi$. Then
\[
  J_b(\pi_S) \le J_S(\pi_S) + d \le J_S(\pi^\star_b) + \epsilon + d \le J_b(\pi^\star_b) + \epsilon + 2d .
\]
For the second statement, fix a policy and its internal randomness. Its normalised cost is a measurable function $f$ of
the normalised path, bounded by $[m, M]$. For any two laws, $|\int f\,dP - \int f\,dQ| \le (M - m)\,\mathrm{TV}(P, Q)$.
Averaging over the policy's randomness and taking the supremum over $\Pi$ gives
$d \le \operatorname{osc}(\tilde J)\,\mathrm{TV}(\tilde P^u_b, \tilde P^u_S)$. The policy influences only the agent's own
state, which is a deterministic function of the path and the policy's randomness, so the distance between path laws
enters once rather than accumulating through policy-dependent state distributions. The distance between path laws is
itself at most the sum of the per-step distances, and so grows with the horizon. $\square$

\paragraph{Proof of \cref{prop:aliasing}.}
Let $A^\star_x(s, \alpha) = Q^\star_x(s, \alpha) - V^\star_x(s) \ge 0$ be the optimal advantage (cost convention) on asset
$x$. By the performance-difference identity, the regret of a policy $\pi$ on $x$ is
$\E_{\pi, x}\big[\sum_k A^\star_x(s_k, a_k)\big]$, which is at least $\E_x[A^\star_x(s_0, a_0)]$ because every term is non-negative. The first state $s_0$ is the
market at $t_0$ together with the initial agent state, so its law does not depend on $\pi$.

A policy acting through $\phi$ draws its first action from a distribution $\pi(\cdot \mid x)$ that depends on the
state only through $x = \phi(s_0)$. For $x \in X$, let $B_a(x)$ be the actions that are at least $\Delta$ worse than
optimal on every state of $a$ with features $x$, and $B_b(x)$ likewise for $b$. By assumption every action lies in
$B_a(x)$ or $B_b(x)$, so $\pi(B_a(x) \mid x) + \pi(B_b(x) \mid x) \ge 1$. Write $\nu_a$ and $\nu_b$ for the laws of
$\phi(s_0)$ under $a$ and $b$, restricted to $X$. Then
\[
  R_a(\pi) \ge \Delta \int_X \pi(B_a(x) \mid x)\, d\nu_a(x) \ge \Delta\rho \int_X \pi(B_a(x) \mid x)\, d\mu(x),
\]
and likewise for $b$, so $R_a(\pi) + R_b(\pi) \ge \Delta\rho \int_X \big[\pi(B_a(x) \mid x) + \pi(B_b(x) \mid x)\big]\,
d\mu(x) \ge \Delta\rho$, and $\max(R_a, R_b) \ge \Delta\rho/2$. The factor $1/2$ cannot be removed for randomised policies:
when one action is optimal on $b$ and exactly $\Delta$ worse on $a$, another the reverse, and nothing else matters, a
uniform mixture of the two has regret $\Delta\rho/2$ on both assets. A
deterministic policy with $X = \{x\}$ chooses one action, which lies in $B_a(x)$ or in $B_b(x)$; its regret on that
asset is at least $\Delta\rho$. $\square$

\paragraph{Proof of \cref{prop:pooled}.}
As $\varepsilon \to 0$, $(1-\varepsilon) F + \varepsilon J_a \to F$ uniformly on the compact $\Theta$, and $\theta'$ is the
unique minimiser of $F$, so every minimiser $\theta_\varepsilon$ converges to $\theta'$; for small $\varepsilon$ it is
interior and solves $G(\theta, \varepsilon) = (1-\varepsilon)\nabla F(\theta) + \varepsilon \nabla J_a(\theta) = 0$. Since
$G(\theta', 0) = 0$ and $\partial_\theta G(\theta', 0) = H$ is invertible, the implicit function theorem gives a unique
continuously differentiable solution $\theta(\varepsilon)$ near $\theta'$, which is therefore $\theta_\varepsilon$, with
derivative $-H^{-1}\partial_\varepsilon G(\theta', 0) = -H^{-1}\big(\nabla J_a(\theta') - \nabla F(\theta')\big) =
-H^{-1} g$ at $\varepsilon = 0$. Hence $\theta_\varepsilon = \theta' - \varepsilon H^{-1} g + o(\varepsilon)$, and a first-order
expansion of $J_a$ at $\theta'$ gives $J_a(\theta_\varepsilon) = J_a(\theta') - \varepsilon\, g^\top H^{-1} g + o(\varepsilon)$.
Adding $J_a(\theta_0) - J_a(\theta')$ gives the decomposition, and $H \succ 0$ gives the bounds. For the last
statement, the descent lemma gives $\min_\Theta J_a \le J_a(\theta' - g/L) \le J_a(\theta') - \|g\|^2/(2L)$, so
$\|g\|^2 \le 2L E_a$, and $g^\top H^{-1} g \le \|g\|^2/\lambda_{\min}(H)$. $\square$ The expansion is the first-order
influence function of an M-estimator~\citep{hampel1974influence, koh2017influence}, applied to the comparator.

In our comparator each test stock contributes about one two-thousandth of the seen policy's episodes (the training
and test-split stocks' 2019 days, size-matched), so the own-data term recovers about $\kappa/1000$ of the excess cost
at most. In control PC1x the gap is visible because $\varepsilon$ is large: the forty test assets share one regime
outside the training range and make up two fifths of the seen policy's data. What remains is the training-mix term, the
difference between pooled policies trained on different mixtures of other stocks, which moves the comparator by about
half a margin. A representation can therefore show a near-zero transfer gap while costing more on every stock than
another. Detecting the first needs a comparator specialised to the stock (large $\varepsilon$), such as
the per-stock policies, whose own data are few here (\cref{app:results}); detecting either needs a comparison of
levels across representations, which our contrasts make. The statement concerns expected costs, smooth (for instance randomised)
policy classes and a comparator that minimises the mixture's cost, as policy-gradient learners such as PPO do. The
tree learner meets none of these assumptions: its policy class is not smooth, and fitted Q-iteration
minimises a regression loss, not the mixture's cost. For it, and for neural fitted Q-iteration, the proposition is a
heuristic: we do not bound their own-data term, which need not be non-negative (we observe small negative gaps), and
their estimation noise adds to both terms. Nor do we estimate $\kappa$ for any comparator used here; for an
overparametrised learner $\lambda_{\min}(H)$ can be close to zero, and the bound is then loose.

%% file: app_controls.tex
\section{Synthetic controls}\label{app:controls}
\input{shared/fig_controls_float}

\paragraph{Market.} The controls use the zero-intelligence order book of \citet{daniels2003quantitative} and
\citet{farmer2005zero} with price-time priority. On each side, limit orders of a fixed size arrive at rate $\alpha$ per
tick at each of the $L = 30$ prices behind the opposite best quote, market orders arrive at rate $\mu$ and sweep the
book, and every resting order is cancelled at rate $\delta$. The simulator emits the same event tables as the ITCH
pipeline, so the replay, the episodes, the learner and the evaluation are the code used on real data.

\paragraph{Exact similarity.} Multiplying $\alpha$, $\mu$ and $\delta$ by $c$ rescales time by $1/c$; multiplying the
order size by $s$ rescales every share count by $s$; multiplying the tick and the price level by $p$ rescales every price
by $p$. Assets related by these maps have identical dynamics in units of mean inter-trade time, mean trade size and mean
spread, which is the premise of \cref{prop:similarity} for \Rms{} units. What the maps leave unchanged are the
dimensionless parameters $(L, \alpha/\mu, \delta/\mu)$. They set the spread in ticks and the queue length in
orders, and the controls vary them on purpose. Kyle--Obizhaeva units are not exact here: $W = P V \sigma$ scales as
$p\,s\,c^{3/2}$ under the maps, so their time unit moves with $p$ and $s$ while the dynamics do not.

\paragraph{Designs.} Every asset draws $c$ log-uniformly in $[0.3, 3]$, $s \in \{1, 3, 10, 30\}$ and
$p \in \{1, 2, 5, 10\}$, so raw units never line up across assets.
\begin{itemize}
  \item \textbf{PC1:} all assets share $\alpha = 0.05$ and $\delta = 0.01$ (per second, per tick).
  \item \textbf{PC1x:} as PC1, but training assets draw $c \in [0.3, 1]$, $s \in \{1, 3\}$ and $p \in \{1, 2\}$,
  and test assets draw $c \in [1.5, 4]$, $s \in \{10, 30\}$ and $p \in \{5, 10\}$.
  \item \textbf{PC2:} $\alpha$ is log-uniform in $[0.02, 1]$, which moves the mean spread from many ticks to one.
  \item \textbf{PC3:} $\delta$ is log-uniform in $[0.002, 0.05]$, which moves the queue length in orders.
\end{itemize}
Each control simulates 60 training assets on two days and 40 test assets on two days, with 150 episodes per asset-day.
Zero-shot policies train on the training assets. Seen policies train on the training assets' first day plus the test
assets' first day, subsampled to the same number of episodes. Both are scored on the test assets' second day, on
identical episodes.

%% file: shared/fig_controls_float.tex
\begin{figure*}[t]
  \centering
  \safegraphic[width=\textwidth]{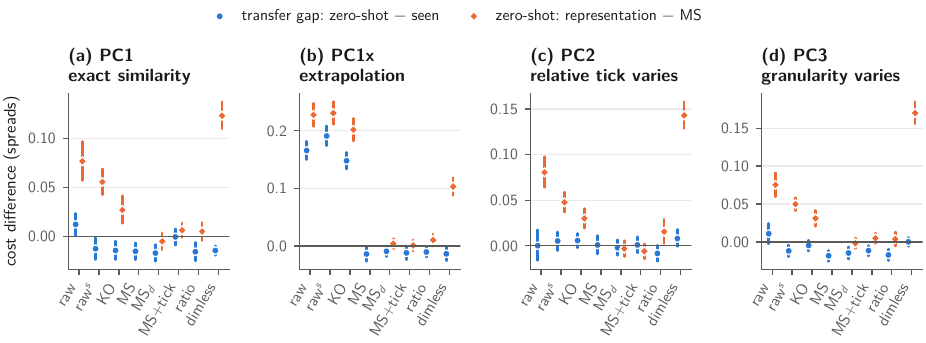}
  \caption{Synthetic controls (zero-intelligence market, advantage learner; 40 test assets, 95\% cluster-bootstrap
  intervals). Circles: transfer gap of each representation, zero-shot minus a policy that also saw the test assets.
  Diamonds: zero-shot cost minus the zero-shot cost in microstructure (\Rms) units. Wrong units cost more in every
  control, but their transfer gap stands out only when the test scales lie outside the training range (b).}
  \label{fig:controls}
\end{figure*}

%% file: app_results.tex
\section{Additional results}\label{app:results}

\subsection{Every representation, by period and tick}\label{app:details}

\begin{figure}[!htbp]
  \centering
  \safegraphic[width=\textwidth]{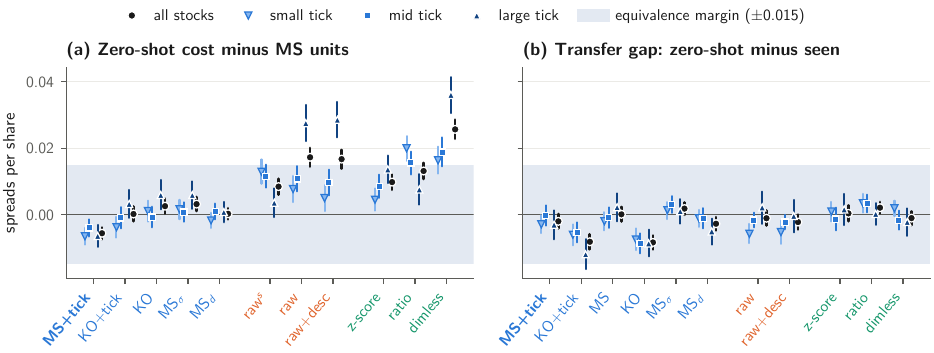}
  \caption{\textbf{Without tick context, losses concentrate where the tick binds.} The 2020--21 contrasts and transfer
  gaps of \cref{fig:transfer} for all stocks and for each tercile of the relative tick. (a) Zero-shot cost minus that
  of \Rms{} units; (b) transfer gap. Three seeds; 95\% stock-bootstrap intervals; band: the margin.}
  \label{fig:transfer_stage1}
\end{figure}

\begin{figure}[!htbp]
  \centering
  \safegraphic[width=\textwidth]{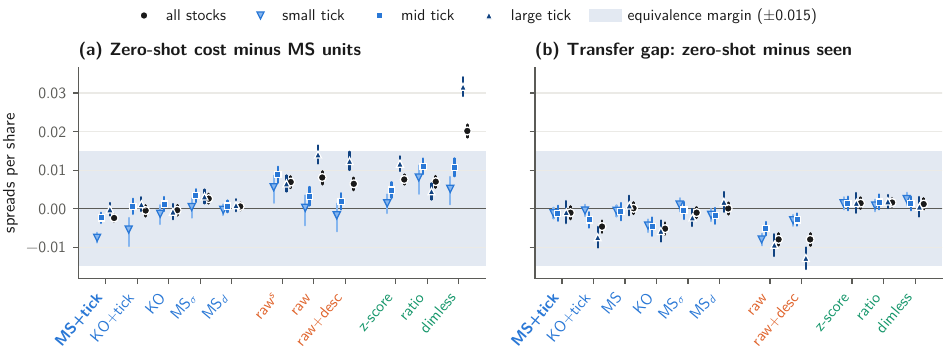}
  \caption{\textbf{Stage~2 (2025--26) by tercile of the relative tick}, as \cref{fig:transfer_stage1}. The contrasts are
  smaller than in 2020--21.}
  \label{fig:transfer_stage2}
\end{figure}

\Cref{tab:stage1,tab:stage2} give the cost of every representation on the two Nasdaq test sets, its difference from
\Rms{} units and its transfer gap. \Cref{fig:transfer_stage1,fig:transfer_stage2} break the contrasts of the two periods down by
tercile of the relative tick.

\begin{table}[!htbp]
\centering
\small
\caption{\textbf{Stage~1} (\EpisodesStageOne{} episodes of \EpisodesStageOneStocks{} stocks, 2020--21). Zero-shot cost
in spreads per share (three seeds), its difference from \Rms{} units, and the transfer gap on the
\RevStageOneTgSeenRThreetickStocks{} stocks whose 2019 days the seen policy saw; 95\% stock-bootstrap intervals. \TabNoteStageOne}
\label{tab:stage1}
\IfFileExists{tab_stage1.tex}{\input{tab_stage1.tex}}{(pending)}
\end{table}

\begin{table}[!htbp]
\centering
\small
\caption{\textbf{Stage~2} (\EpisodesStageTwo{} episodes of \EpisodesStageTwoStocks{} stocks, 2025--26), as
\cref{tab:stage1}; transfer gaps on the \RevStageTwoTgSeenRThreetickStocks{} stocks the seen policy saw. \TabNoteStageTwo}
\label{tab:stage2}
\IfFileExists{tab_stage2.tex}{\input{tab_stage2.tex}}{(pending)}
\end{table}

\subsection{Robustness checks and the designs behind them}\label{app:robust}

The checks below are exploratory; costs are gross of fees, in spreads per share.

\paragraph{Fill rules, latency and windows (validation).}
The exact fill rule of \cref{sec:setup} lies between two bounds: a pessimistic rule fills a resting order only when
the market trades through its price or an opposite order crosses it, and an optimistic rule fills it as if it were at
the front of its queue. Raw units cost $\RobpessimisticContrastsRZeroRThreeAll$ more than \Rms{} units under the
pessimistic rule, $\RoboptimisticContrastsRZeroRThreeAll$ under the optimistic one and $\ValContrastsRZeroRThreeAll$
under the exact one, and the learned \Rmst{} policy's gain over posting and waiting ranges from
$\RoboptimisticVsPostThenCrossRThreetickAll$ to $\RobpessimisticVsPostThenCrossRThreetickAll$. A latency of zero,
1\,ms or 10\,ms instead of 100\,\textmu s changes no contrast in the third decimal, because decisions are seconds apart
(raw units: $\RoblatOneZeromsContrastsRZeroRThreeAll$ at 10\,ms). Units computed from a trailing window of 15 or 60
minutes instead of 30, with the tasks unchanged, change the cost of the \Rmst{} policies by
$\WinFifteenMstickZeroVsReg$ and $\WinSixtyMstickZeroVsReg$; policies retrained with them cost
$\WinFifteenMstickRetrainVsReg$ and $\WinSixtyMstickRetrainVsReg$ relative to the original ones, and raw units still
lose $\WinFifteenRawVsRetrainMstick$ and $\WinSixtyRawVsRetrainMstick$ to them.

\paragraph{Seeds, days and training samples.}
A seed changes only the learner's row and feature subsampling on a fixed training set. On the 2020--21 stocks every
comparison keeps its sign and its standing against the margin when seeds are resampled along with stocks,
and again when days are; the standard deviation of a contrast across seeds is at most 0.0025. To measure how
much the training sample matters, five representations and the seen \Rmst{} policy were refitted on \BootRefits{}
resamples of the training stocks drawn with replacement and scored on validation. Across refits (and across seeds),
the standard deviation is $\BootValHOneSd$ ($\BootValHOneSeedSd$) for the transfer gap, $\BootValHTwoSd$
($\BootValHTwoSeedSd$) for raw units' loss, $\BootValHThreeSd$ ($\BootValHThreeSeedSd$) for z-scores' loss and
$\BootValHSixSd$ ($\BootValHSixSeedSd$) for the gain from tick context. How raw units' loss divides between the cost
scale and the features varies more: raw features with the cost in spreads minus \Rms{} has a standard deviation of
$\BootValHEightSd$ ($\BootValHEightSeedSd$), and the cost scale's share of the loss is \BootValCostUnitShare{} across
refits against \BootValCostUnitShareSeed{} on the fixed sample (\PpoCostUnitShare{} with PPO on validation and
\RevStageOneNnCostUnitShare{} with the neural learner on the 2020--21 stocks).

We widened the test-set intervals of \BootChecked{} comparisons, six per period, by this variation: the transfer gap
of \Rmst, the losses of raw units and z-scores, the gain from tick context, and the two parts of raw units' loss
(raw features with the cost in spreads against \Rms{} units, which should be within the margin if the cost scale were
all, and the cost in cents against the cost in spreads). Widened by the whole refit variance, which counts the seed
variation twice because a refit also changes the seed, \BootKeptFull{} keep their conclusion: the transfer gap and
z-scores' loss in both periods, and raw units' loss in 2020--21 (in 2025--26 its interval is
$[\BootStageTwoHTwoFullWideLo, \BootStageTwoHTwoFullWideHi]$). Widened only by the variance in excess of the seeds',
a rule we adopted after the first, \BootKept{} keep: raw units' 2025--26 loss, the gain from tick context in both
periods (only because its refit standard deviation is below its seed standard deviation) and the cost in cents in
2020--21 join them. That raw features with the cost in spreads come within the margin of \Rms{} units fails under
both rules, as does the cost in cents in 2025--26. The transfer gap stays well inside the margin either way
($[\BootStageOneHOneFullWideLo, \BootStageOneHOneFullWideHi]$ in 2020--21).

\paragraph{Small orders and the cost distribution.}
A non-intrusive replay is most accurate for orders small relative to the book. On the \SmallOrderShare{} of
validation episodes whose order is at most a quarter of the window's mean touch depth (\SmallOrderSmallEpisodes{}
episodes), the contrasts keep their signs, with intervals that include zero: raw units $\SmallOrderSmallHTwo$
($[\SmallOrderSmallHTwoLo, \SmallOrderSmallHTwoHi]$, against $\SmallOrderLargeHTwo$ for larger orders), z-scores
$\SmallOrderSmallHThree$ ($\SmallOrderLargeHThree$) and tick context $\SmallOrderSmallHSix$ ($\SmallOrderLargeHSix$).
Over the whole cost distribution, with episodes weighted equally, the zero-shot \Rmst{} policy's median cost is the
same as the seen policy's or lower ($\DistStageOneHOneMed$ in 2020--21, $\DistStageTwoHOneMed$ in 2025--26) and its
worst 5\% of episodes cost less (by $\DistStageOneHOneCvarAbs$ and $\DistStageTwoHOneCvarAbs$). Raw minus \Rms{} is
$\DistStageOneHTwoMean$ in the mean in 2020--21 but $\DistStageOneHTwoMed$ ($[\DistStageOneHTwoMedLo,
\DistStageOneHTwoMedHi]$) at the median and $\DistStageOneHTwoCvar$ in the mean of the worst 5\% of episodes; in
2025--26, $\DistStageTwoHTwoMean$, $\DistStageTwoHTwoMed$ and $\DistStageTwoHTwoCvar$.

\paragraph{The stock's own data.}
Four comparators give the zero-shot \Rmst{} policy what a per-stock calibration would use. \emph{History}: \Rmst{} plus four features of the stock's 2019 history, the current window's time
between trades, trade size and spread relative to their 2019 medians and the tick relative to the 2019 spread; it is
scored on the \RevStageOneTgSeenRThreetickStocks{} 2020--21 test stocks that traded in 2019, and likewise in
2025--26. \emph{Fill profile}: \Rmst{} plus four per-stock constants, the mean cost on the stock's 2019 days of
four simple rules (posting and waiting, waiting and crossing at the eighth decision, improving the quote by a tick,
and a queue rule) relative to crossing at once, each shrunk toward the training stocks' mean by empirical Bayes.
\emph{Per-stock corrections}: offsets of the pooled advantages of each action, the stock's shrunken mean fitted-Q
residuals, with the shrinkage chosen by replaying a left-out day. \emph{Per-stock training}: the same learner trained
on one stock's five 2019 days, and the queue rule (join while the shares ahead are few relative to the time left and
the book's imbalance is not adverse, otherwise cross) tuned per stock. The last three are scored on the validation
stocks' held-out days. The placebo gives every validation episode the units of a random other stock on the same day,
leaving the task, the path and the tick unchanged; policies that use no window units are unaffected by it to the last
digit.

\paragraph{Outside the training range.}
Every representation was also trained only on the training stocks whose activity, mean trade size or mean spread lies
in the middle half of its range, and scored on 2020--21 stocks outside that range against the same representation
trained on as many episodes from all training stocks (its extrapolation gap). The training range spans a factor of
\XtrapDescActivityRangeRatio{} in activity, and the stocks outside it a factor of \XtrapDescActivityOutsideRatio.

\paragraph{The RL-free diagnostic.}
$\Dhat$ (\cref{prop:regret}) replays 22 simple rules, crossing at one of four decisions, posting and waiting,
improving and then crossing, and sixteen queue rules, on each stock's 2019 days only, and compares their normalised
costs with those on the training stocks. Its median over validation stocks is $\DhatLevelRThree$ in \Rms{} units,
$\DhatLevelRTwo$ in \Rko{} units and $\DhatLevelRZero$ in raw units: simple rules cost most alike across stocks in
microstructure units. Its rank correlation with the per-stock loss when the spread lies outside the training range,
on validation stocks, is $\XtrapSpreadRThreetickDhatrho$ for \Rmst{} and $\XtrapSpreadRZeroDhatrho$ for raw units.

\paragraph{Training on the test period.}
To separate the age of the training data from the stock, the \Rmst{} and raw-unit policies were also trained, by the
same recipe and size-matched, on the training stocks' own days of each test period and scored once on it. The
zero-shot \Rmst{} policy minus this contemporaneous one is $\CtmpStageOneMstickYears$ in 2020--21 and
$\CtmpStageTwoMstickYears$ ($[\CtmpStageTwoMstickYearsLo, \CtmpStageTwoMstickYearsHi]$) in 2025--26; trained on the
test period, raw units still lose $\CtmpStageOneRawVsMstick$ and $\CtmpStageTwoRawVsMstick$ to \Rmst.

\subsection{What the units supply}\label{app:supply}

\paragraph{Information and a cost scale.}
Raw features with the cost in spreads (\Rrawr) and raw features plus every window statistic the units are built from
(\Rrawi) separate the parts of raw units' loss (\cref{fig:supply}a). With trees on the 2020--21 stocks, raw units
lose $\InfoStageOneTreesRawVsMstick$ to \Rmst, raw features with the cost in spreads $\InfoStageOneTreesRawrVsMstick$,
and raw features with the window statistics as well $\InfoStageOneTreesInfoVsMstick$; with the neural learner,
$\InfoStageOneNnRawVsMstick$, $\InfoStageOneNnRawrVsMstick$ and $\InfoStageOneNnInfoVsMstick$. \Cref{tab:rev-arms}
gives the neural learner's contrasts and the learned alternatives with seeds resampled, and \cref{tab:rev-econ} the
size of the main contrasts in basis points, gross and net of fees.

\paragraph{Data.}
Averaged over \CurveSeeds{} seeds, a pooled \Rmst{} policy trained on a thousand episodes of other stocks
($\CurveMsTickOneK$ spreads) already does better than the same learner trained on the stock's own five days
(\cref{fig:supply}b). Raw units need roughly ten times the episodes to reach the same cost: \Rmst{} on ten thousand
episodes costs $\CurveMsTickTenK$ and raw units on a hundred thousand $\CurveRawHundredK$. Given the window statistics
and the cost in spreads, raw features need about three times the data of \Rmst{} when episodes are few
($\CurveInfoInfoOneK$ on a thousand) and match it from thirty thousand on ($\CurveInfoInfoThirtyK$ against
$\CurveMsTickThirtyK$).

\begin{table}[!htbp]
\centering
\small
\caption{\textbf{Learners and learned normalisation.} Paired cost differences in spreads per share (crypto: volatility
per trade), seeds averaged, with 95\% intervals that resample stocks and training seeds together. Learned scale: each
training stock's reward divided by a scale learned from its own data (PopArt-style); random rescaling: training prices
and sizes rescaled at random over a factor of 100; descriptors: the four inputs that \Rrawd{} adds.}
\label{tab:rev-arms}
\IfFileExists{tab_arms_learners.tex}{\input{tab_arms_learners.tex}}{(pending)}
\end{table}

\begin{table}[!htbp]
\centering
\small
\caption{\textbf{Economic size.} Paired cost differences in basis points of the price: gross, net of Nasdaq's fees,
and with stocks weighted by notional instead of equally; 95\% stock-bootstrap intervals.}
\label{tab:rev-econ}
\IfFileExists{tab_revision_econ.tex}{\input{tab_revision_econ.tex}}{(pending)}
\end{table}

\subsection{IEX}\label{app:iex}

The IEX test covers \AnalysedIexTest{} episodes of \AnalysedIexTestStocks{} stocks on one-tick books and
\AnalysedIexTestWide{} episodes of \AnalysedIexTestWideStocks{} stocks on wider books, on the ten days that followed
the freezing of the policies. IEX delays every incoming order by 350\,$\mu$s, much of its trading is against hidden
orders, and its visible book is often thin. The comparators are fitted Q-iteration and PPO trained by the zero-shot
policies' recipe on 386{,}618 one-tick episodes of five earlier IEX days.

\paragraph{One-tick books.}
With every episode weighted equally, zero-shot fitted Q-iteration and PPO beat crossing at once
($\IexTestEpisodeIOneAll$ and $\IexTestPpoEpisodePOneAll$) and raw units, and are within the margin of 0.020 of the
same learners trained on IEX (fitted Q-iteration $\IexTestEpisodeITwoAll$, 95\% CI $[\IexReITwoEpisodeLoNinetyfive,
\IexReITwoEpisodeHiNinetyfive]$; PPO $\IexTestLmEpisodePFourAll$, $[\IexRePFourEpisodeLoNinetyfive,
\IexRePFourEpisodeHiNinetyfive]$). PPO's gap would not fit within the Nasdaq margin of 0.015. PPO trained on Nasdaq's
one-tick books alone recovers about a quarter of it, so most of it is what IEX's own 2025--26 data add; fine-tuning
the zero-shot policy on IEX's training episodes recovers all of it ($\FtTestFtVsZeroEpisode$, fine-tuned minus
zero-shot).

\paragraph{One episode, and the two averages.}
With every stock weighted equally, a stock carries the same weight however few episodes it has, so the average is
worth only \NeffIexTest{} equally weighted episodes of \AnalysedIexTest{} (\NeffIexTestShare), against
\NeffStageOneShare{} and \NeffStageTwoShare{} on the Nasdaq test sets. \IexDiagStock{} had one episode, a
\IexDiagSide{} of \IexDiagQ{} shares on a book whose spread averaged \IexDiagSpreadTicks{} ticks. Crossing at once
cost \IexDiagImmediate{} spreads; the zero-shot policy waited, other traders took the bid, and at the forced cross
IEX's remaining bids lay 16\% below the market, where the replay sold at \IexDiagCostZero{} spreads per share. A real
order would have been routed to other venues, and price bands would have stopped the trade, so the episode shows a
limit of the replay, not of transfer. It weighs \IexDiagWeight{} times an average episode, and without it the
stock-weighted comparisons of fitted Q-iteration take the signs of the episode-weighted ones (with the IEX-trained
policy $\IexNoNhpITwo$, $[\IexNoNhpITwoLo, \IexNoNhpITwoHi]$). The median over stocks and a cap of 1.33 spreads on each
episode's cost point the same way.

\paragraph{Wider books.}
On books wider than one tick for most of the window, raw units, z-scores and raw units with a learned per-stock reward
scale all lose to \Rmst{} with either average ($\IexWideWFiveAll$ to $\IexWideWFourAll$ with stocks weighted equally,
$\IexTestWideEpisodeWFiveAll$ to $\IexTestWideEpisodeWFourAll$ with episodes). These are the books on which the
replay is least realistic, because crosses walk into IEX's sparse visible depth, which a routed order would not reach;
they test how a policy copes with an unfamiliar, thinner book, not what trading on IEX costs.

%% file: tab_stage1.tex
\begin{tabular}{@{}lrll@{}}
\toprule
Representation & Cost & Minus \Rms & Transfer gap \\
\midrule
\multicolumn{4}{@{}l}{\emph{Invariant units}} \\
\quad\Rms & 0.439 & \emph{reference} & $+0.000$ {\scriptsize[$-0.002$, $+0.002$]} \\
\quad\Rmst & 0.432 & $-0.006$ {\scriptsize[$-0.007$, $-0.004$]} & $-0.002$ {\scriptsize[$-0.004$, $+0.000$]} \\
\quad\Rmss & 0.442 & $+0.003$ {\scriptsize[$+0.001$, $+0.005$]} & $+0.002$ {\scriptsize[$-0.000$, $+0.004$]} \\
\quad\Rmsd & 0.438 & $+0.000$ {\scriptsize[$-0.001$, $+0.002$]} & $-0.003$ {\scriptsize[$-0.005$, $-0.001$]} \\
\quad\Rko & 0.440 & $+0.003$ {\scriptsize[$+0.000$, $+0.005$]} & $-0.008$ {\scriptsize[$-0.010$, $-0.006$]} \\
\quad\Rkot & 0.437 & $+0.000$ {\scriptsize[$-0.002$, $+0.002$]} & $-0.008$ {\scriptsize[$-0.010$, $-0.006$]} \\
\addlinespace
\multicolumn{4}{@{}l}{\emph{Raw units}} \\
\quad\Rraw & 0.455 & $+0.017$ {\scriptsize[$+0.014$, $+0.020$]} & $-0.001$ {\scriptsize[$-0.003$, $+0.001$]} \\
\quad\Rrawd & 0.455 & $+0.017$ {\scriptsize[$+0.014$, $+0.019$]} & $-0.002$ {\scriptsize[$-0.004$, $+0.000$]} \\
\quad\Rrawr & 0.445 & $+0.008$ {\scriptsize[$+0.006$, $+0.011$]} & n/a \\
\addlinespace
\multicolumn{4}{@{}l}{\emph{Standard normalisations}} \\
\quad\Rz & 0.447 & $+0.010$ {\scriptsize[$+0.007$, $+0.012$]} & $+0.000$ {\scriptsize[$-0.002$, $+0.003$]} \\
\quad\Rratio & 0.451 & $+0.013$ {\scriptsize[$+0.011$, $+0.016$]} & $+0.002$ {\scriptsize[$+0.000$, $+0.004$]} \\
\quad\Rdimless & 0.463 & $+0.026$ {\scriptsize[$+0.023$, $+0.029$]} & $-0.001$ {\scriptsize[$-0.003$, $+0.001$]} \\
\bottomrule
\end{tabular}

%% file: tab_stage2.tex
\begin{tabular}{@{}lrll@{}}
\toprule
Representation & Cost & Minus \Rms & Transfer gap \\
\midrule
\multicolumn{4}{@{}l}{\emph{Invariant units}} \\
\quad\Rms & 0.477 & \emph{reference} & $+0.000$ {\scriptsize[$-0.001$, $+0.002$]} \\
\quad\Rmst & 0.474 & $-0.002$ {\scriptsize[$-0.003$, $-0.001$]} & $-0.001$ {\scriptsize[$-0.003$, $+0.001$]} \\
\quad\Rmss & 0.479 & $+0.003$ {\scriptsize[$+0.001$, $+0.004$]} & $-0.001$ {\scriptsize[$-0.002$, $+0.000$]} \\
\quad\Rmsd & 0.477 & $+0.001$ {\scriptsize[$-0.000$, $+0.002$]} & $+0.000$ {\scriptsize[$-0.001$, $+0.002$]} \\
\quad\Rko & 0.476 & $-0.000$ {\scriptsize[$-0.002$, $+0.001$]} & $-0.005$ {\scriptsize[$-0.007$, $-0.004$]} \\
\quad\Rkot & 0.476 & $-0.000$ {\scriptsize[$-0.002$, $+0.001$]} & $-0.005$ {\scriptsize[$-0.006$, $-0.003$]} \\
\addlinespace
\multicolumn{4}{@{}l}{\emph{Raw units}} \\
\quad\Rraw & 0.485 & $+0.008$ {\scriptsize[$+0.006$, $+0.010$]} & $-0.008$ {\scriptsize[$-0.010$, $-0.006$]} \\
\quad\Rrawd & 0.483 & $+0.006$ {\scriptsize[$+0.005$, $+0.008$]} & $-0.008$ {\scriptsize[$-0.010$, $-0.006$]} \\
\quad\Rrawr & 0.483 & $+0.007$ {\scriptsize[$+0.005$, $+0.008$]} & n/a \\
\addlinespace
\multicolumn{4}{@{}l}{\emph{Standard normalisations}} \\
\quad\Rz & 0.484 & $+0.008$ {\scriptsize[$+0.006$, $+0.009$]} & $+0.001$ {\scriptsize[$+0.000$, $+0.003$]} \\
\quad\Rratio & 0.484 & $+0.007$ {\scriptsize[$+0.005$, $+0.008$]} & $+0.002$ {\scriptsize[$+0.001$, $+0.003$]} \\
\quad\Rdimless & 0.498 & $+0.020$ {\scriptsize[$+0.019$, $+0.022$]} & $+0.001$ {\scriptsize[$-0.000$, $+0.003$]} \\
\bottomrule
\end{tabular}

%% file: tab_arms_learners.tex
\begin{tabular}{@{}lllll@{}}
\toprule
Contrast & Val. & Stage 1 & Stage 2 & Crypto \\
\midrule
\multicolumn{5}{@{}l}{\emph{Neural fitted Q-iteration}} \\
\quad \Rraw{} $-$ \Rms & \begin{tabular}[t]{@{}l@{}}$+0.042$\\{\scriptsize[$+0.029$, $+0.056$]}\end{tabular} & \begin{tabular}[t]{@{}l@{}}$+0.031$\\{\scriptsize[$+0.023$, $+0.038$]}\end{tabular} & \begin{tabular}[t]{@{}l@{}}$+0.020$\\{\scriptsize[$+0.013$, $+0.027$]}\end{tabular} & \begin{tabular}[t]{@{}l@{}}$+0.064$\\{\scriptsize[$-0.020$, $+0.161$]}\end{tabular} \\
\quad \Rrawd{} $-$ \Rms & \begin{tabular}[t]{@{}l@{}}$+0.033$\\{\scriptsize[$+0.023$, $+0.043$]}\end{tabular} & \begin{tabular}[t]{@{}l@{}}$+0.026$\\{\scriptsize[$+0.019$, $+0.032$]}\end{tabular} & \begin{tabular}[t]{@{}l@{}}$+0.017$\\{\scriptsize[$+0.010$, $+0.025$]}\end{tabular} & \begin{tabular}[t]{@{}l@{}}$+0.029$\\{\scriptsize[$-0.034$, $+0.101$]}\end{tabular} \\
\quad \Rz{} $-$ \Rmst & \begin{tabular}[t]{@{}l@{}}$+0.025$\\{\scriptsize[$+0.014$, $+0.037$]}\end{tabular} & \begin{tabular}[t]{@{}l@{}}$+0.021$\\{\scriptsize[$+0.014$, $+0.028$]}\end{tabular} & \begin{tabular}[t]{@{}l@{}}$+0.011$\\{\scriptsize[$+0.005$, $+0.018$]}\end{tabular} & \begin{tabular}[t]{@{}l@{}}$+0.122$\\{\scriptsize[$+0.070$, $+0.188$]}\end{tabular} \\
\quad \Rko{} $-$ \Rms & \begin{tabular}[t]{@{}l@{}}$-0.002$\\{\scriptsize[$-0.010$, $+0.006$]}\end{tabular} & \begin{tabular}[t]{@{}l@{}}$-0.001$\\{\scriptsize[$-0.005$, $+0.004$]}\end{tabular} & \begin{tabular}[t]{@{}l@{}}$-0.002$\\{\scriptsize[$-0.006$, $+0.002$]}\end{tabular} & \begin{tabular}[t]{@{}l@{}}$-0.025$\\{\scriptsize[$-0.078$, $+0.025$]}\end{tabular} \\
\quad \Rmst{} $-$ \Rms & \begin{tabular}[t]{@{}l@{}}$-0.007$\\{\scriptsize[$-0.016$, $+0.001$]}\end{tabular} & \begin{tabular}[t]{@{}l@{}}$-0.008$\\{\scriptsize[$-0.014$, $-0.002$]}\end{tabular} & \begin{tabular}[t]{@{}l@{}}$-0.005$\\{\scriptsize[$-0.011$, $+0.001$]}\end{tabular} & \begin{tabular}[t]{@{}l@{}}$-0.001$\\{\scriptsize[$-0.042$, $+0.043$]}\end{tabular} \\
\quad \Rrawr{} $-$ \Rms & \begin{tabular}[t]{@{}l@{}}$+0.005$\\{\scriptsize[$-0.004$, $+0.014$]}\end{tabular} & \begin{tabular}[t]{@{}l@{}}$+0.004$\\{\scriptsize[$-0.002$, $+0.010$]}\end{tabular} & \begin{tabular}[t]{@{}l@{}}$-0.000$\\{\scriptsize[$-0.006$, $+0.005$]}\end{tabular} & \begin{tabular}[t]{@{}l@{}}$+0.011$\\{\scriptsize[$-0.045$, $+0.072$]}\end{tabular} \\
\quad \Rraw{} $-$ \Rrawr & \begin{tabular}[t]{@{}l@{}}$+0.037$\\{\scriptsize[$+0.023$, $+0.051$]}\end{tabular} & \begin{tabular}[t]{@{}l@{}}$+0.027$\\{\scriptsize[$+0.020$, $+0.034$]}\end{tabular} & \begin{tabular}[t]{@{}l@{}}$+0.020$\\{\scriptsize[$+0.014$, $+0.027$]}\end{tabular} & \begin{tabular}[t]{@{}l@{}}$+0.053$\\{\scriptsize[$-0.028$, $+0.155$]}\end{tabular} \\
\quad \Rraw, learned scale $-$ \Rmst & \begin{tabular}[t]{@{}l@{}}$+0.007$\\{\scriptsize[$-0.002$, $+0.015$]}\end{tabular} & \begin{tabular}[t]{@{}l@{}}$+0.005$\\{\scriptsize[$-0.000$, $+0.010$]}\end{tabular} & \begin{tabular}[t]{@{}l@{}}$-0.001$\\{\scriptsize[$-0.007$, $+0.004$]}\end{tabular} & \begin{tabular}[t]{@{}l@{}}$+0.003$\\{\scriptsize[$-0.069$, $+0.068$]}\end{tabular} \\
\quad \Rraw, random rescaling $-$ \Rmst & \begin{tabular}[t]{@{}l@{}}$+0.054$\\{\scriptsize[$+0.043$, $+0.065$]}\end{tabular} & \begin{tabular}[t]{@{}l@{}}$+0.044$\\{\scriptsize[$+0.037$, $+0.053$]}\end{tabular} & \begin{tabular}[t]{@{}l@{}}$+0.031$\\{\scriptsize[$+0.025$, $+0.038$]}\end{tabular} & \begin{tabular}[t]{@{}l@{}}$+0.084$\\{\scriptsize[$+0.006$, $+0.171$]}\end{tabular} \\
\quad \Rmst, descriptors $-$ \Rmst & \begin{tabular}[t]{@{}l@{}}$-0.003$\\{\scriptsize[$-0.011$, $+0.004$]}\end{tabular} & \begin{tabular}[t]{@{}l@{}}$-0.001$\\{\scriptsize[$-0.006$, $+0.005$]}\end{tabular} & \begin{tabular}[t]{@{}l@{}}$-0.001$\\{\scriptsize[$-0.007$, $+0.004$]}\end{tabular} & \begin{tabular}[t]{@{}l@{}}$+0.001$\\{\scriptsize[$-0.052$, $+0.046$]}\end{tabular} \\
\addlinespace
\multicolumn{5}{@{}l}{\emph{Fitted Q-iteration with trees}} \\
\quad \Rmst, descriptors $-$ \Rmst & \begin{tabular}[t]{@{}l@{}}$+0.001$\\{\scriptsize[$-0.004$, $+0.006$]}\end{tabular} & \begin{tabular}[t]{@{}l@{}}$+0.002$\\{\scriptsize[$-0.001$, $+0.004$]}\end{tabular} & \begin{tabular}[t]{@{}l@{}}$+0.000$\\{\scriptsize[$-0.001$, $+0.002$]}\end{tabular} & \begin{tabular}[t]{@{}l@{}}$-0.004$\\{\scriptsize[$-0.039$, $+0.031$]}\end{tabular} \\
\quad \Rraw, learned scale $-$ \Rms & \begin{tabular}[t]{@{}l@{}}$+0.006$\\{\scriptsize[$-0.001$, $+0.014$]}\end{tabular} & \begin{tabular}[t]{@{}l@{}}$+0.006$\\{\scriptsize[$+0.003$, $+0.009$]}\end{tabular} & \begin{tabular}[t]{@{}l@{}}$+0.003$\\{\scriptsize[$+0.001$, $+0.005$]}\end{tabular} & \begin{tabular}[t]{@{}l@{}}$+0.054$\\{\scriptsize[$-0.017$, $+0.133$]}\end{tabular} \\
\quad \Rraw, learned scale $-$ \Rrawr & \begin{tabular}[t]{@{}l@{}}$+0.001$\\{\scriptsize[$-0.005$, $+0.006$]}\end{tabular} & \begin{tabular}[t]{@{}l@{}}$-0.002$\\{\scriptsize[$-0.004$, $+0.001$]}\end{tabular} & \begin{tabular}[t]{@{}l@{}}$-0.004$\\{\scriptsize[$-0.005$, $-0.002$]}\end{tabular} & \begin{tabular}[t]{@{}l@{}}$+0.003$\\{\scriptsize[$-0.027$, $+0.041$]}\end{tabular} \\
\addlinespace
\multicolumn{5}{@{}l}{\emph{Neural minus trees}} \\
\quad \Rmst & \begin{tabular}[t]{@{}l@{}}$-0.003$\\{\scriptsize[$-0.010$, $+0.004$]}\end{tabular} & \begin{tabular}[t]{@{}l@{}}$+0.000$\\{\scriptsize[$-0.004$, $+0.005$]}\end{tabular} & \begin{tabular}[t]{@{}l@{}}$+0.002$\\{\scriptsize[$-0.003$, $+0.007$]}\end{tabular} & \begin{tabular}[t]{@{}l@{}}$+0.020$\\{\scriptsize[$-0.030$, $+0.076$]}\end{tabular} \\
\quad \Rraw & \begin{tabular}[t]{@{}l@{}}$+0.023$\\{\scriptsize[$+0.011$, $+0.037$]}\end{tabular} & \begin{tabular}[t]{@{}l@{}}$+0.017$\\{\scriptsize[$+0.011$, $+0.024$]}\end{tabular} & \begin{tabular}[t]{@{}l@{}}$+0.016$\\{\scriptsize[$+0.011$, $+0.022$]}\end{tabular} & \begin{tabular}[t]{@{}l@{}}$+0.058$\\{\scriptsize[$+0.008$, $+0.125$]}\end{tabular} \\
\bottomrule
\end{tabular}

%% file: tab_revision_econ.tex
\begin{tabular}{@{}llll@{}}
\toprule
Contrast & Gross & Net of fees & Notional-weighted \\
\midrule
\multicolumn{4}{@{}l}{\emph{Validation (2019)}} \\
\quad \Rraw{} $-$ \Rms & $+0.23$ {\scriptsize[$+0.13$, $+0.33$]} & $-0.07$ {\scriptsize[$-0.18$, $+0.04$]} & $+0.23$ {\scriptsize[$+0.10$, $+0.39$]} \\
\quad \Rrawr{} $-$ \Rms & $+0.16$ {\scriptsize[$+0.09$, $+0.24$]} & $+0.17$ {\scriptsize[$+0.09$, $+0.25$]} & $+0.20$ {\scriptsize[$+0.09$, $+0.31$]} \\
\quad \Rraw{} $-$ \Rrawr & $+0.07$ {\scriptsize[$-0.01$, $+0.15$]} & $-0.24$ {\scriptsize[$-0.32$, $-0.15$]} & $+0.04$ {\scriptsize[$-0.03$, $+0.11$]} \\
\quad \Rmst{} $-$ \Rms & $-0.06$ {\scriptsize[$-0.11$, $-0.01$]} & $+0.01$ {\scriptsize[$-0.04$, $+0.05$]} & $-0.03$ {\scriptsize[$-0.09$, $+0.02$]} \\
\quad \Rmst{} $-$ post, then cross & $-1.01$ {\scriptsize[$-1.17$, $-0.86$]} & $-0.77$ {\scriptsize[$-0.94$, $-0.60$]} & $-0.97$ {\scriptsize[$-1.16$, $-0.78$]} \\
\quad \Rmst{} $-$ cross at once & $-2.94$ {\scriptsize[$-3.22$, $-2.66$]} & $-3.58$ {\scriptsize[$-3.89$, $-3.28$]} & $-2.43$ {\scriptsize[$-2.80$, $-2.11$]} \\
\addlinespace
\multicolumn{4}{@{}l}{\emph{Stage 1 (2020--21)}} \\
\quad \Rraw{} $-$ \Rms & $+0.46$ {\scriptsize[$+0.35$, $+0.58$]} & $-0.02$ {\scriptsize[$-0.14$, $+0.10$]} & $+0.18$ {\scriptsize[$+0.14$, $+0.23$]} \\
\quad \Rrawr{} $-$ \Rms & $+0.41$ {\scriptsize[$+0.31$, $+0.52$]} & $+0.38$ {\scriptsize[$+0.27$, $+0.49$]} & $+0.15$ {\scriptsize[$+0.10$, $+0.19$]} \\
\quad \Rraw{} $-$ \Rrawr & $+0.05$ {\scriptsize[$-0.03$, $+0.13$]} & $-0.40$ {\scriptsize[$-0.48$, $-0.32$]} & $+0.04$ {\scriptsize[$+0.01$, $+0.07$]} \\
\quad \Rmst{} $-$ \Rms & $-0.14$ {\scriptsize[$-0.20$, $-0.09$]} & $-0.06$ {\scriptsize[$-0.11$, $+0.00$]} & $-0.07$ {\scriptsize[$-0.09$, $-0.05$]} \\
\quad \Rmst{} $-$ post, then cross & $-2.36$ {\scriptsize[$-2.61$, $-2.14$]} & $-2.05$ {\scriptsize[$-2.30$, $-1.82$]} & $-1.07$ {\scriptsize[$-1.14$, $-1.00$]} \\
\quad \Rmst{} $-$ cross at once & $-6.11$ {\scriptsize[$-6.47$, $-5.77$]} & $-7.25$ {\scriptsize[$-7.63$, $-6.89$]} & $-2.95$ {\scriptsize[$-3.11$, $-2.79$]} \\
\addlinespace
\multicolumn{4}{@{}l}{\emph{Stage 2 (2025--26)}} \\
\quad \Rraw{} $-$ \Rms & $+0.21$ {\scriptsize[$+0.02$, $+0.41$]} & $-0.33$ {\scriptsize[$-0.53$, $-0.14$]} & $+0.15$ {\scriptsize[$+0.11$, $+0.18$]} \\
\quad \Rrawr{} $-$ \Rms & $+0.40$ {\scriptsize[$+0.21$, $+0.59$]} & $+0.45$ {\scriptsize[$+0.26$, $+0.64$]} & $+0.19$ {\scriptsize[$+0.15$, $+0.22$]} \\
\quad \Rraw{} $-$ \Rrawr & $-0.19$ {\scriptsize[$-0.29$, $-0.09$]} & $-0.78$ {\scriptsize[$-0.88$, $-0.68$]} & $-0.04$ {\scriptsize[$-0.06$, $-0.01$]} \\
\quad \Rmst{} $-$ \Rms & $-0.32$ {\scriptsize[$-0.44$, $-0.21$]} & $-0.20$ {\scriptsize[$-0.33$, $-0.10$]} & $-0.10$ {\scriptsize[$-0.11$, $-0.08$]} \\
\quad \Rmst{} $-$ post, then cross & $-3.08$ {\scriptsize[$-3.36$, $-2.81$]} & $-2.65$ {\scriptsize[$-2.93$, $-2.37$]} & $-1.24$ {\scriptsize[$-1.30$, $-1.18$]} \\
\quad \Rmst{} $-$ cross at once & $-8.92$ {\scriptsize[$-9.46$, $-8.41$]} & $-10.23$ {\scriptsize[$-10.78$, $-9.70$]} & $-4.03$ {\scriptsize[$-4.21$, $-3.86$]} \\
\bottomrule
\end{tabular}

%% file: app_pendulum.tex
\section{A second domain: pendulums}
\subsection{A family of pendulums}\label{app:pendulum}

This section asks the questions of the synthetic controls (\cref{app:controls}) on a task family outside finance in
which similarity is exact by construction (exploratory).

\paragraph{Systems.} A point mass $m$ on a rod of length $l$ under gravity $g$, with viscous damping $b$ and a torque
limit $u_{\max}$, is upright at $\theta = 0$: $\ddot\theta = (g/l)\sin\theta - b\,\dot\theta/(ml^2) + u/(ml^2)$,
integrated with semi-implicit Euler at step $\Delta t$. Its natural units are the time $\tau_0 = \sqrt{l/g}$, the
angular velocity $1/\tau_0$ and the torque $mgl$. Every dimensionless group is fixed ($u_{\max}/mgl = \PendURatio$, so
the pendulum must be pumped up; $b\tau_0/ml^2 = \PendDamping$; $\Delta t/\tau_0 = \PendDtRatio$; \PendHorizon{} steps),
so a policy on natural-unit features is exactly invariant (\cref{prop:similarity}, with nothing approximate). Systems
draw $m$ and $l$ log-uniformly in $[0.3, 3]$ and $g$ in $[3, 30]$, so $\tau_0 \in [0.1, 1]$; out-of-range test systems
then have $l$ and $g$ moved, at fixed $lg$, to $\tau_0$ in $[3, 9]$ or $[1/90, 1/30]$. The actions are the torques
$\{-1, -\frac12, 0, \frac12, 1\}\,u_{\max}$. A step costs $\tilde c = (1 - \cos\theta) + 0.1\,(\omega\tau_0)^2 +
0.01\,(u/mgl)^2$ in natural units and $mgl\,\tilde c$, an energy, in raw units (spreads against cents). A policy is
scored by its mean $\tilde c$ per step on \PendEpisodes{} episodes per test system from the hanging position.
Representations: natural units, $(\cos\theta, \sin\theta, \omega\tau_0)$, cost $\tilde c$; \Rraw, $(\cos\theta,
\sin\theta, \omega)$ with $\omega$ in rad/s, cost $mgl\,\tilde c$; \Rrawr, the same with cost $\tilde c$; \Rrawi,
\Rrawr{} plus $\tau_0$, so that the natural angular velocity is a product of two inputs; \Rrawd, \Rraw{} plus
$(m, l, g)$.

\paragraph{Learners and design.} Discounted fitted Q-iteration ($\gamma = \PendGamma$, \PendIters{} iterations) on
branched data, \PendStates{} uniform random states per system ($\omega\tau_0 \in [-8, 8]$) with all five actions each,
regresses advantages as our main learner does: each iteration's targets are split into their mean over actions,
fitted on the state, and the deviations from it, fitted on state and action. Trees use the default settings
(\cref{tab:learners}) with \PendRounds{} rounds. The neural learner uses MLPs with \PendNnLayers{}
hidden layers of \PendNnHidden{} \PendNnAct{} units on the inputs of the paper's neural learner (Adam, learning rate
\PendNnLr, \PendNnEpochs{} epochs per iteration, warm-started). The time step, the state sampling, the torque grid, the
data size, the rounds and the neural settings were set in pilots in natural units, for a reliable swing-up, before any
other representation was fitted; at $\Delta t/\tau_0 = 0.05$, as first planned, even exact value iteration truncated at
\PendIters{} iterations is upright for only \PendGridFineUp{} of the last quarter of an episode. Zero-shot policies
train on \PendTrainSystems{} systems. (a) A seen policy also sees a group of \PendGroup{} in-range test systems, each
1/72 of its data; (b) the same out of range; (c) in a sweep, each of $G$ in-range test systems is a share $\varepsilon$
of a seen policy's data, $(\varepsilon, G) = (1/64, 4), (1/16, 4), (1/4, 2), (1/2, 1)$, on \PendSweepSystems{} systems
per seed spanning its range of $\tau_0$; (d) a pooled seen policy sees all \PendOutSystems{} out-of-range systems, each
1/32 of its data. Each of \PendSeeds{} seeds draws
\PendTestSystems{} in-range and \PendOutSystems{} out-of-range test systems, the data and the learner's randomness. We also fit
trees with four times the data (\PendXfourStates{} states per system) and, for the sweep only, use seeds up to
\PendSweepSeeds{} in all. Intervals are 95\% bootstrap intervals over test systems,
stratified by seed; in the sweep they also resample seeds, and so fits.

\paragraph{A small transfer gap is not evidence of invariance.} Natural units cost \PendInRangeCostNat{} per step in
range and \PendOutRangeCostNat{} out of range. With trees, \Rraw{} costs \PendInRangeRawVsNat{} more in range
[\PendInRangeRawVsNatLo, \PendInRangeRawVsNatHi], mostly on fast systems (\PendInRangeTauLowRawVsNat{} on the third with
the smallest $\tau_0$, \PendInRangeTauHighRawVsNat{} on the largest), yet its transfer gap is \PendInRangeTgRaw{}
[\PendInRangeTgRawLo, \PendInRangeTgRawHi]; \Rrawd{} never swings up (\PendInRangeRawdVsNat{} more) and has a gap of
\PendInRangeTgRawd. The neural learner agrees (\Rraw: \PendNnInRangeRawVsNat{} more, gap \PendNnInRangeTgRaw). Out of
range every representation but natural units fails (trees, \Rraw: \PendOutRangeRawVsNat{} more), and the gaps do not
show it (\PendOutRangeTgRaw), because the seen policy, in the same representation, fails the same way. That is aliasing
(\cref{prop:aliasing}), not a small share: in raw features the states that decide a swing-up, slow motion near the top
and the bottom, look alike for every system, so no seen policy in those features can tell the new systems from the
training ones, whatever $\varepsilon$ is. Design (d) shows it: with the out-of-range systems at half the data, the gap
of \Rraw{} stays at \PendOutPoolTgRaw{} with trees and \PendNnOutPoolTgRaw{} with the neural learner, while that of
\Rrawi, whose features separate the systems, reaches \PendNnOutPoolTgRawi{} with the neural learner (trees:
\PendOutPoolTgRawi).

\paragraph{The share $\varepsilon$.} Intervals over test systems alone hold the fits fixed; with them, the transfer
gap of natural units, which is training noise, excludes zero at \PendSweepNatExclZeroOneLevel{} of the four shares.
With \PendSweepSeeds{} seeds and intervals that also resample seeds, \PendSweepNatExclZero{} does. The gap of \Rraw{}
minus that of natural units is \PendSweepDiffSixtyFourth{} [\PendSweepDiffSixtyFourthLo, \PendSweepDiffSixtyFourthHi],
\PendSweepDiffSixteenth{} and \PendSweepDiffQuarter{} at $\varepsilon = 1/64, 1/16, 1/4$, and \PendSweepDiffHalf{}
[\PendSweepDiffHalfLo, \PendSweepDiffHalfHi] at $1/2$, against a zero-shot excess of \PendSweepExcessRaw; from 1/64 to
1/2 it rises by \PendSweepHalfMinusSixtyFourthDiff{} [\PendSweepHalfMinusSixtyFourthDiffLo,
\PendSweepHalfMinusSixtyFourthDiffHi]. The sweep resolves a step, not a dose-response: a seen policy for which the test
system is half the data recovers part of the excess, but no growth is resolved between 1/64 and 1/4, the shares a
transfer gap uses. It is consistent with a gap of order $\varepsilon$ (\cref{prop:pooled})
but is not evidence for that rate.

\paragraph{Information versus division.} With trees, \Rrawi{} costs \PendInRangeRawiVsNat{} more than natural units in
range [\PendInRangeRawiVsNatLo, \PendInRangeRawiVsNatHi], no less than \Rrawr{} (\PendInRangeRawiVsRawr), and comes
close only in the middle third of $\tau_0$ (\PendInRangeTauMidRawiVsNat). Four times the data makes it worse: the gap
grows by \PendXfourInRangeRawiGapChange{} [\PendXfourInRangeRawiGapChangeLo, \PendXfourInRangeRawiGapChangeHi] while
natural units are unchanged (\PendXfourInRangeNatChange). A shortage of splits to approximate the product
$\omega\tau_0$ would shrink with more data, so this points instead to an instability of fitted Q-iteration with
trees on these inputs, which we have not diagnosed. The neural learner comes close in range: \PendNnInRangeRawiVsNat{}
[\PendNnInRangeRawiVsNatLo, \PendNnInRangeRawiVsNatHi] more than natural units, \PendNnInRangeRawiVsRawr{} relative to
\Rrawr, \PendNnInRangeTauLowRawiVsNat{} on the fastest third; out of range it fails like the others
(\PendNnOutRangeRawiVsNat). The real-data pattern, close within the training range and behind outside it, holds for the
neural learner; the in-range failure belongs to the tree learner and says nothing about information against
division.

\paragraph{Caveats.} (i) Outside the sweep, intervals hold the trained policies fixed, and the transfer gap of natural
units is training noise: with trees \PendInRangeTgNat{} in range, from \PendInRangeTgNatFitMin{} to
\PendInRangeTgNatFitMax{} across the twelve seen policies; with the neural learner \PendNnInRangeTgNat, from
\PendNnInRangeTgNatFitMin{} to \PendNnInRangeTgNatFitMax. Gaps of that size are noise even when their intervals exclude
zero. (ii) Out-of-range seen policies pool four similar systems per group. (iii) \PendInRangeOutsideSample{} of the
\PendInRangeSystemsTotal{} in-range test systems lie slightly outside the range of $\tau_0$ spanned by their seed's
training systems. (iv) The neural learner costs more than trees in natural units (\PendNnInRangeCostNat{} against
\PendInRangeCostNat), and its GPU training is not bit-for-bit reproducible.

\begin{figure}[h]
  \centering
  \safegraphic[width=\textwidth]{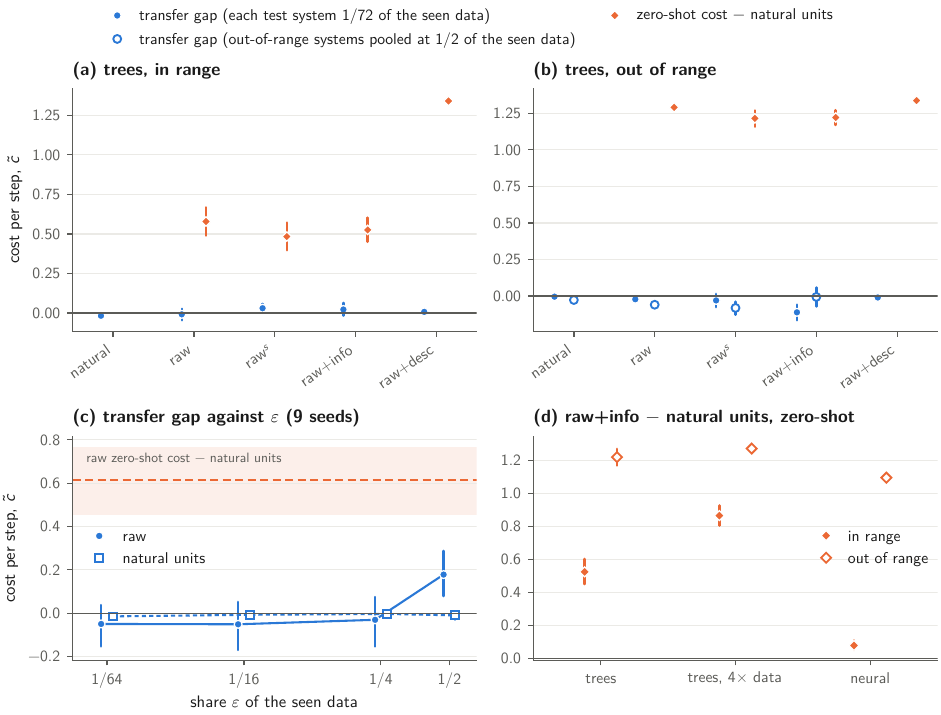}
  \caption{The pendulum family (exploratory). (a, b) Trees: transfer gap (zero-shot minus seen, blue) and zero-shot
  cost minus that of natural units (orange) for each representation, in and out of range; open circles in (b) are
  seen policies with all out-of-range systems pooled at half their data (design (d)). Natural units
  cost \PendInRangeCostNat{} per step. (c) Trees: transfer gap against the test system's share $\varepsilon$ of the
  seen policy's data (log scale), for \Rraw{} and natural units, over \PendSweepSeeds{} seeds; the dashed line and band
  are \Rraw's zero-shot excess over natural units on the same systems. (d) \Rrawi{} minus natural units, zero-shot, in
  range (filled) and out of range (open), for trees, trees with four times the data, and the neural learner. 95\%
  bootstrap intervals over test systems, stratified by seed; in (c) they also resample seeds.}
  \label{fig:pendulum}
\end{figure}

%% file: app_carl.tex
\subsection{CARL's pendulum, without exact similarity}\label{app:carl}

Natural units are exact in the family above by construction. This subsection asks the same questions on the
contextual pendulum of CARL~\citep{benjamins2023contextualize}, where they are not, with its held-out-context evaluation.

\paragraph{Environment and contexts.} We re-implemented gym's Pendulum-v1 as CARL embeds it, from gym's source
(neither package was run): $\omega' = \operatorname{clip}(\omega + (\frac{3g}{2l}\sin\theta + \frac{3}{ml^2}u)\Delta t,
\pm 8\,\mathrm{rad/s})$, $\theta' = \theta + \omega'\Delta t$, $|u| \le 2$\,N\,m, cost $\theta^2 + 0.1\,\omega^2 +
0.001\,u^2$ per step, \CarlHorizon{} steps from $\theta \sim U[-\pi, \pi]$, $\omega \sim U[-1, 1]$; defaults $g = 10$,
$l = m = 1$, $\Delta t = 0.05$\,s. As in CARL, $(g, l, m)$ varies and the time step and the speed and torque limits
stay fixed in SI units. In natural units (time $\tau_0 = \sqrt{2l/3g}$, torque $mgl/2$) the problem, cost included,
depends on $\delta = \Delta t/\tau_0$ and $\bar u = u_{\max}/(mgl/2)$ (the speed limit $\omega_{\max}\tau_0$ follows
from $\delta$); at the default $\delta = \CarlDefaultDelta$, $\bar u = \CarlDefaultUbar$, $\omega_{\max}\tau_0 =
\CarlDefaultWmax$. The groups vary across contexts, so natural units are only approximately invariant, as the
one-cent tick makes the units approximate for stocks (\cref{sec:theory}). The torque is discretised to $\{-1,
-\frac12, 0, \frac12, 1\} \times 2$\,N\,m. Per seed, \CarlTrainContexts{} training and \CarlTestContexts{} held-out
in-range test contexts draw $g$, $l$, $m$ log-uniformly in $[\CarlBoxLo, \CarlBoxHi]$ times the default. Two
out-of-range sets of \CarlSimContexts{} put every feature outside that box, as in the extrapolation mode of the
protocol CARL uses~\citep{kirk2023generalisation}: \emph{context out, problem in range} applies $(g, l, m) \to (kg,
kl, m/k^2)$, which leaves gym's dynamics unchanged, until each feature is at least 1.25 times outside the box;
\emph{every feature out} draws $g$ and $m$ \CarlOutFactorLo--\CarlOutFactorHi{} times below the box and $l$ as far
above it (slower, lighter pendulums: $\tau_0$ \CarlExtTauMin--\CarlExtTauMax\,s against
\CarlTrainTauMin--\CarlTrainTauMax\,s in training, so $\delta$ is out of range).

\paragraph{Range.} The range of contexts was set on nat+ctx alone. With gym's limits fixed, fast pendulums with
little torque cannot be swung up at all, so the box keeps contexts where an oracle (grid value iteration with the
context known) solves at least 90\% of episodes on \CarlCheckBoxSolvable{} of contexts, and the out-of-range sets
extrapolate only towards slower, lighter pendulums. The learners keep the settings of the family above (\CarlStates{}
states per context, $\omega$ uniform on $[-8, 8]$\,rad/s).

\paragraph{Representations and design.} \Rraw: gym's $(\cos\theta, \sin\theta, \omega)$; raw+ctx: \Rraw{} plus $(g,
l, m)$, CARL's concatenated context; \Rrawi: \Rraw{} plus $\tau_0$; nat: $(\cos\theta, \sin\theta, \omega\tau_0)$;
nat+ctx: nat plus $\log\delta$, $\log\bar u$, $\log\omega_{\max}\tau_0$, the analogue of \Rmst. \Rrawi{} has the
information of nat, raw+ctx that of nat+ctx. All train and are scored on gym's cost (per step, \CarlEpisodes{}
episodes per test context, the same draws for every policy). Seen policies, as above: one per group of
\CarlGroup{} in-range test contexts, each 1/72 of the data; one pooling all out-of-range contexts, each 1/32.
\CarlSeeds{} seeds, \CarlFitsTr{} fits with trees and \CarlFitsNn{} with the neural learner; 95\% intervals resample
seeds, then contexts.

\paragraph{(L1) A small transfer gap is not evidence here either.} In range the gaps are small next to the
differences in cost: with trees, \Rraw{} costs $\CarlTrInRawVsNatc$ per step more than nat+ctx
($[\CarlTrInRawVsNatcLo, \CarlTrInRawVsNatcHi]$) with a gap of $\CarlTrInTgRaw$ ($[\CarlTrInTgRawLo,
\CarlTrInTgRawHi]$), raw+ctx $\CarlTrInRawcVsNatc$ with $\CarlTrInTgRawc$; with the neural learner, nat costs
$\CarlNnInNatVsNatc$ more with a gap of $\CarlNnInTgNat$, and \Rraw's gap ($\CarlNnInTgRaw$, $[\CarlNnInTgRawLo, \CarlNnInTgRawHi]$) is a quarter of its
excess ($\CarlNnInRawVsNatc$): well above nat+ctx's gap ($\CarlNnInTgNatc$), it flags raw units here but shows only
part of what they lose. Per seed the gap is \CarlNnInTgShareRawSeedMin{} to \CarlNnInTgShareRawSeedMax{} of \Rraw's
excess with the neural learner and \CarlTrInTgShareRawSeedMin{} to \CarlTrInTgShareRawSeedMax{} with trees. Out of range, where the features cannot tell the new contexts from the old, aliasing
(\cref{prop:aliasing}) hides the loss from a comparator that saw them at half its data: on contexts whose problem is
in range, neural \Rraw{} costs $\CarlNnSimRawVsNatc$ more than nat+ctx with a gap of $\CarlNnSimTgRaw$, and with trees
that comparator does no better than the zero-shot policy ($\CarlTrSimTgRaw$, $[\CarlTrSimTgRawLo,
\CarlTrSimTgRawHi]$). Where the features do tell them apart, the gaps show the failure (raw+ctx there:
$\CarlTrSimTgRawc$ with trees, $\CarlNnSimTgRawc$ neural).

\paragraph{(L2) Information comes close in range, with the neural learner only.} Neural raw+ctx is within noise of
nat+ctx in range ($\CarlNnInRawcVsNatc$, $[\CarlNnInRawcVsNatcLo, \CarlNnInRawcVsNatcHi]$), and \Rrawi{} of nat
($\CarlNnInRawiVsNat$). Out of range they are not: raw+ctx costs $\CarlNnSimRawcVsNatc$ more
($[\CarlNnSimRawcVsNatcLo, \CarlNnSimRawcVsNatcHi]$) on contexts whose problem is in range, where \Rrawi, its
$\tau_0$ in range, still matches nat ($\CarlNnSimRawiVsNat$); with every feature out, raw+ctx costs
$\CarlNnExtRawcVsNatc$ more than nat+ctx and \Rrawi{} $\CarlNnExtRawiVsNat$ more than nat. Trees come close in
neither (raw+ctx $\CarlTrInRawcVsNatc$, \Rrawi{} $\CarlTrInRawiVsNat$ against nat in range, positive on every seed),
as in the family above, where more data made them worse: the failure belongs to the tree learner, not to the
information.

\paragraph{(L3) The dimensionless context helps one learner, in range.} Neural nat+ctx minus nat is
$\CarlNnInNatcVsNat$ in range ($[\CarlNnInNatcVsNatLo, \CarlNnInNatcVsNatHi]$; per seed $\CarlNnInNatcVsNatSeedMin$
to $\CarlNnInNatcVsNatSeedMax$) and $\CarlNnSimNatcVsNat$ on contexts whose problem is in range, as the tick context
helps on stocks. With every feature out, where the groups themselves extrapolate, it is $\CarlNnExtNatcVsNat$:
nat+ctx balances in no episode (share \CarlNnExtSolvedNatc), nat in all. With trees, which hold predictions constant
beyond the last split, the groups cost a little everywhere: $\CarlTrInNatcVsNat$ in range ($[\CarlTrInNatcVsNatLo,
\CarlTrInNatcVsNatHi]$; per seed $\CarlTrInNatcVsNatSeedMin$ to $\CarlTrInNatcVsNatSeedMax$), $\CarlTrExtNatcVsNat$
with every feature out. Similarity is close enough here for trees: nat alone costs $\CarlTrInNatVsOracle$ more than
the oracle in range (\CarlOracleInCost) and $\CarlTrExtNatVsOracle$ with every feature out. Why the extra inputs
cost trees we have not established.

\paragraph{Caveats.} (i) One environment, re-implemented, with a discretised torque, contexts narrowed to where the
task is solvable, and extrapolation in one direction. (ii) Three seeds make the intervals rough; nat+ctx's gap,
mostly training noise, ranges over the in-range seen policies from $\CarlTrInTgNatcFitMin$ to $\CarlTrInTgNatcFitMax$
(trees) and $\CarlNnInTgNatcFitMin$ to $\CarlNnInTgNatcFitMax$ (neural). (iii) The neural learner balances less
reliably than trees (nat+ctx solves \CarlNnInSolvedNatc{} of in-range episodes, against \CarlTrInSolvedNatc) at a similar cost (\CarlNnInCostNatc, \CarlTrInCostNatc), and is not bit-for-bit
reproducible on GPU. (iv) The range was chosen on nat+ctx, whose out-of-range costs were printed while choosing it.

\begin{figure}[!htbp]
  \centering
  \safegraphic[width=\textwidth]{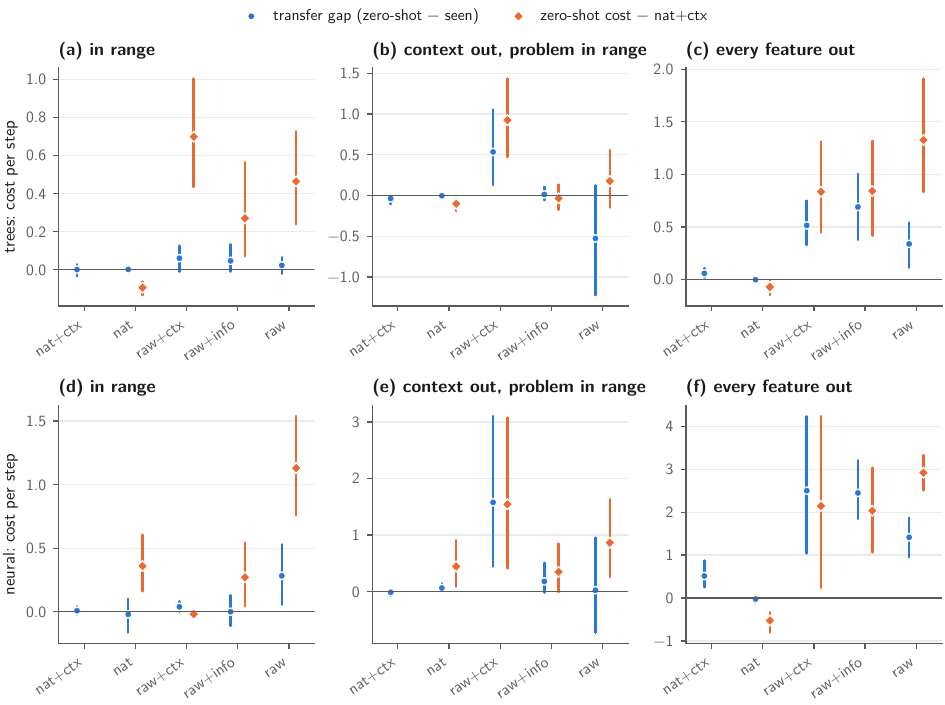}
  \caption{CARL's pendulum (exploratory). Transfer gap (zero-shot minus seen, blue) and zero-shot cost minus that of
  nat+ctx (orange) per representation; top trees, bottom the neural learner. (a, d) In-range test contexts, each 1/72
  of a seen policy's data; (b, e) contexts whose $(g, l, m)$ lie out of range but whose problem does not; (c, f)
  every feature out of range (slower, lighter pendulums); out of range the seen policy pools all such contexts, each
  1/32 of its data. Gym's cost per step; in range nat+ctx costs \CarlTrInCostNatc{} (trees) and \CarlNnInCostNatc{}
  (neural), the oracle \CarlOracleInCost. 95\% bootstrap intervals over seeds, then contexts.}
  \label{fig:carl}
\end{figure}